\pdfoutput=1

\documentclass[11pt]{article}

\usepackage[preprint]{acl}

\usepackage{times}
\usepackage{latexsym}

\usepackage[T1]{fontenc}

\usepackage[utf8]{inputenc}

\usepackage{microtype}

\usepackage{inconsolata}

\usepackage{hyperref}
\usepackage{url}
\usepackage{amsthm}

\usepackage{mdframed}
\usepackage{framed}
\usepackage{colortbl}
\usepackage{graphicx}
\usepackage{tabularx} 
\usepackage{booktabs}
\usepackage{makecell}
\usepackage{lipsum} 
\usepackage{siunitx} %
\usepackage{tikz}

\usepackage{multirow}
\usepackage{multicol}
\usepackage{graphicx}
\usepackage{adjustbox}
\usepackage[normalem]{ulem}
\usepackage{epigraph}
\usepackage{color,soul}
\usepackage{amssymb} 

\usepackage{subcaption}
\usepackage{amsmath}
\usepackage{cleveref}
\usepackage{tabularx}

\usepackage{listings}
\usepackage{xcolor}
\usepackage{longtable}

\colorlet{punct}{red!60!black}
\definecolor{bg}{HTML}{EEEEEE}
\definecolor{delim}{RGB}{20,105,176}
\colorlet{numb}{magenta!60!black}

\lstdefinelanguage{json}{
    basicstyle=\normalfont\ttfamily,
    numbers=left,
    numberstyle=\scriptsize,
    stepnumber=1,
    numbersep=8pt,
    showstringspaces=false,
    breaklines=true,
    frame=lines,
    backgroundcolor=\color{bg},
    literate=
     *{0}{{{\color{numb}0}}}{1}
      {1}{{{\color{numb}1}}}{1}
      {2}{{{\color{numb}2}}}{1}
      {3}{{{\color{numb}3}}}{1}
      {4}{{{\color{numb}4}}}{1}
      {5}{{{\color{numb}5}}}{1}
      {6}{{{\color{numb}6}}}{1}
      {7}{{{\color{numb}7}}}{1}
      {8}{{{\color{numb}8}}}{1}
      {9}{{{\color{numb}9}}}{1}
      {:}{{{\color{punct}{:}}}}{1}
      {,}{{{\color{punct}{,}}}}{1}
      {\{}{{{\color{delim}{\{}}}}{1}
      {\}}{{{\color{delim}{\}}}}}{1}
      {[}{{{\color{delim}{[}}}}{1}
      {]}{{{\color{delim}{]}}}}{1},
}

\definecolor{shadecolor}{RGB}{245,245,245}
\definecolor{rowcolor}{rgb}{0.85, 0.85, 0.88}
\definecolor{rowcolor1}{RGB}{243,248,240} %
\definecolor{rowcolor2}{RGB}{250,239,239} %
\definecolor{highlightcolor}{rgb}{1, 0, 0} %
\definecolor{keybgcolor}{rgb}{1, 1, 0.8} %
\definecolor{my_green}{RGB}{141,185,111} %
\definecolor{my_red}{RGB}{208,100,100} %

\newcommand\keyword[1]{%
\tikz[baseline=(word.base)]{
  \node[rounded corners=1pt,fill=rowcolor,anchor=base, inner sep=1.2pt] (word) {#1};}%
}

\newcommand{\redunline}[1]{\textcolor{my_red}{#1}}

\newcommand{\greenunline}[1]{\textcolor{my_green}{#1}}

\definecolor{trashbin_blue}{RGB}{23,211,253}
\definecolor{hydrant_green}{RGB}{37,253,53}
\definecolor{bench_orange}{RGB}{255,195,45}

\setul{0.5ex}{0.3ex}

\definecolor{red}{RGB}{208,100,100}
\definecolor{green}{RGB}{141,185,111}
\definecolor{blue}{RGB}{95,148,207}
\definecolor{gray}{RGB}{149,149,149}
\definecolor{purple}{RGB}{120,100,179}

\newcommand{\theme}[1]{\setulcolor{gray}\ul{#1}}
\newcommand{\background}[1]{\setulcolor{red}\ul{#1}}
\newcommand{\persona}[1]{\setulcolor{green}\ul{#1}}
\newcommand{\plot}[1]{\setulcolor{blue}\ul{#1}}

\newcommand{\themeprompt}{\color{gray}{\#\#\# Theme} \\ \vspace{-1em} \color{gray}\{theme\}\\}
\newcommand{\bgprompt}{\color{red}{\#\#\# Background} \\ \vspace{-1em} \color{red}\{background\}\\}
\newcommand{\personaprompt}{\color{green}{\#\#\# Persona} \\\vspace{-1em} \color{green}\{persona\}\\}
\newcommand{\eventprompt}{\color{blue}{\#\#\# Event} \\\vspace{-1em} \color{blue}\{event\}\\}
\newcommand{\finalendingprompt}{\color{blue}{\#\#\# Ending} \\\vspace{-1em} \color{blue}\{ending\}\\}
\newcommand{\twistprompt}{\color{blue}{\#\#\# Twist} \\\vspace{-1em} \color{blue}\{twist\}\\}

\usepackage{fancyhdr}
\title{MoPS: Modular Story Premise Synthesis \\ for Open-Ended Automatic Story Generation}

\author{
Yan Ma\textsuperscript{\rm{1,3,4}}\footnotemark[1] \quad
\textbf{Yu Qiao}\textsuperscript{\rm{3}} \quad
\textbf{Pengfei Liu}\textsuperscript{\rm{2,3,4}}\footnotemark[2]\\
\textsuperscript{1}Fudan University \
\textsuperscript{2}Shanghai Jiao Tong University \\
\textsuperscript{3}Shanghai AI Laboratory \
\textsuperscript{4}Generative AI Research Lab (GAIR) \\
\texttt{yanma23@m.fudan.edu.cn} \quad 
\texttt{qiaoyu@pjlab.org.cn} \quad
\texttt{pengfei@sjtu.edu.cn}
}

\begin{document}
\maketitle
\renewcommand{\thefootnote}{\fnsymbol{footnote}}
\footnotetext[1]{\,This work was done during Yan Ma's internship at Shanghai AI Laboratory.}
\footnotetext[2]{\,Corresponding author.}
\renewcommand{\thefootnote}{\arabic{footnote}}

\begin{abstract}
A story premise succinctly defines a story's main idea, foundation, and trajectory.
It serves as the initial trigger in automatic story generation.
Existing sources of story premises are limited by a lack of diversity, uneven quality, and high costs that make them difficult to scale.
In response, we introduce \textbf{Mo}dular Story \textbf{P}remise \textbf{S}ynthesis (MoPS) which breaks down story premises into modules like background and persona for automated design and generation.
MoPS consists of three phases: (1) Pre-collect a consistent set of candidates for each module to form a nested dictionary. (2) Extract a key path from the nested dictionary as the premise design. (3) Instruct an LLM to integrate the design into a coherent premise sentence.
Thorough evaluations demonstrate that our synthesized premises excel in diversity, fascination, completeness, and originality compared to those induced from large language models and captured from public story datasets.
Similarly, the extended novels and scripts generated from our premises also exhibit higher quality.
In supplementary materials, we provide the MoPS code suite, along with 7.6k generated premises and 1k extended stories.
Code: \url{https://github.com/GAIR-NLP/MoPS}.
\end{abstract}

\section{Introduction}
\label{sec:intro}
\vspace{-0.03in}
\begin{quote}
``\textit{If a story is going to fail, \\it will do so first at the premise level.}'' \\
-- \underline{Anatomy of a Premise Line}
\end{quote}

\begin{table*}[t]
    \footnotesize
    \centering
    \begin{tabularx}{\textwidth}{>{\columncolor{rowcolor1}}X|>{\columncolor{rowcolor2}}X} 
    \toprule
    {\keyword{\textbf{Premise A:}} \textit{\small A powerful Roman general,\textcolor{my_green}{\uline{granted immortality by a vengeful deity}}, must choose between loyalty to the empire and leading a rebellion, ultimately decide to bring justice and equality to the corrupt rulers of the Roman Empire. }} & {\keyword{\textbf{Premise B:}} \textit{A Roman general chooses between loyalty to empire and leading a rebellion for justice and equality. }} \\

    \midrule

    \vspace{0.1cm}
    \keyword{\textbf{Poster A:}} 
    \qquad\quad\adjustbox{width=0.15\textwidth,valign=c}{\includegraphics{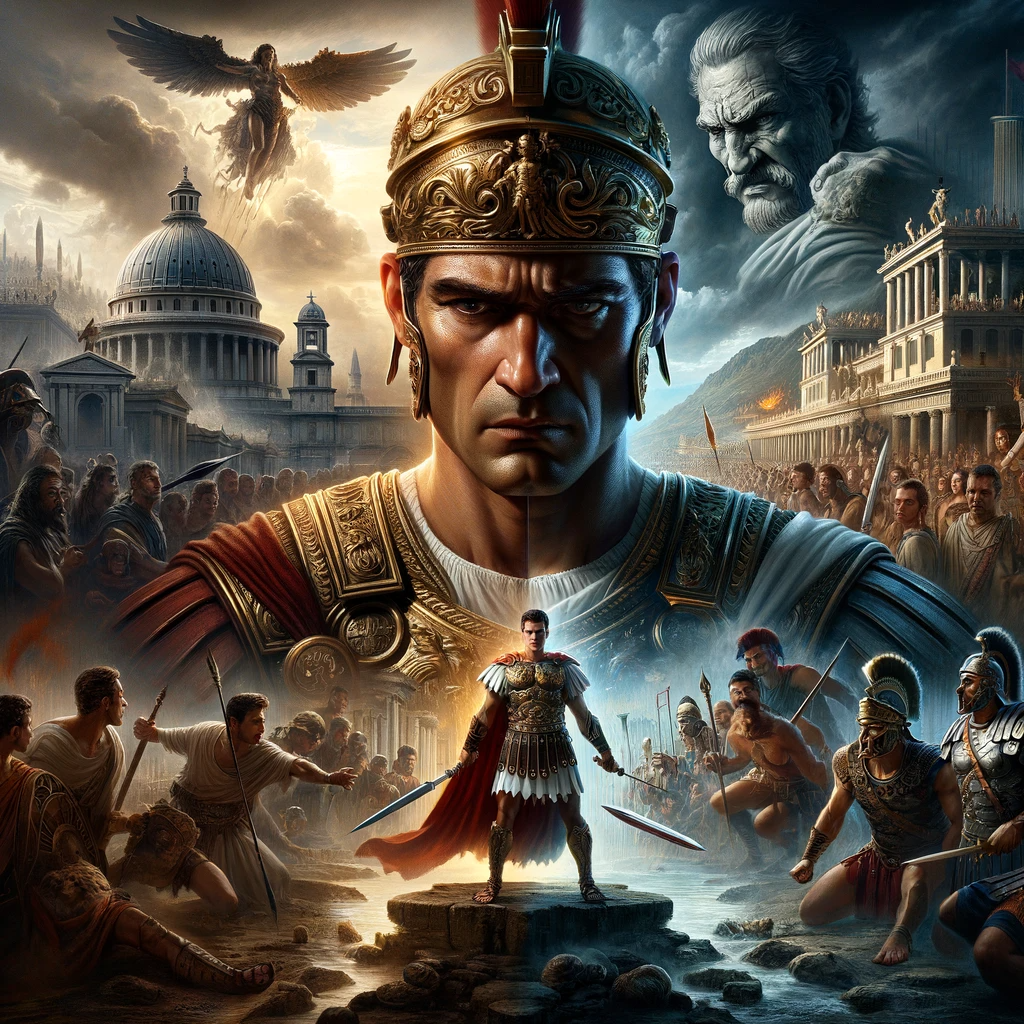}}
    \newline
    \vspace{0.2cm}
    
    \keyword{\textbf{Story A:}} Centuries ago, in the heart of ancient Rome, there was a powerful general named Lucius, \greenunline{who had been granted immortality by a vengeful deity}. As the empire crumbled under the weight of corruption and tyranny, Lucius faced a fateful decision: remain loyal to the empire that had once revered him, or lead a rebellion to bring justice and equality to its corrupt rulers. After witnessing the suffering of the oppressed, Lucius chose to forsake his loyalty to the empire, \greenunline{vowing to use his immortality to dismantle the corrupt regime}. With unwavering determination, he rallied the oppressed citizens, becoming a symbol of hope and change. Together, they waged a relentless battle for a brighter future, where justice and equality would reign, and the legacy of Lucius would live on through the ages as \greenunline{a beacon of righteous rebellion against tyranny}.
    &
    \vspace{0.1cm}
    \keyword{\textbf{Poster B:}} 
    \qquad\quad\adjustbox{width=0.15\textwidth,valign=c}{\includegraphics{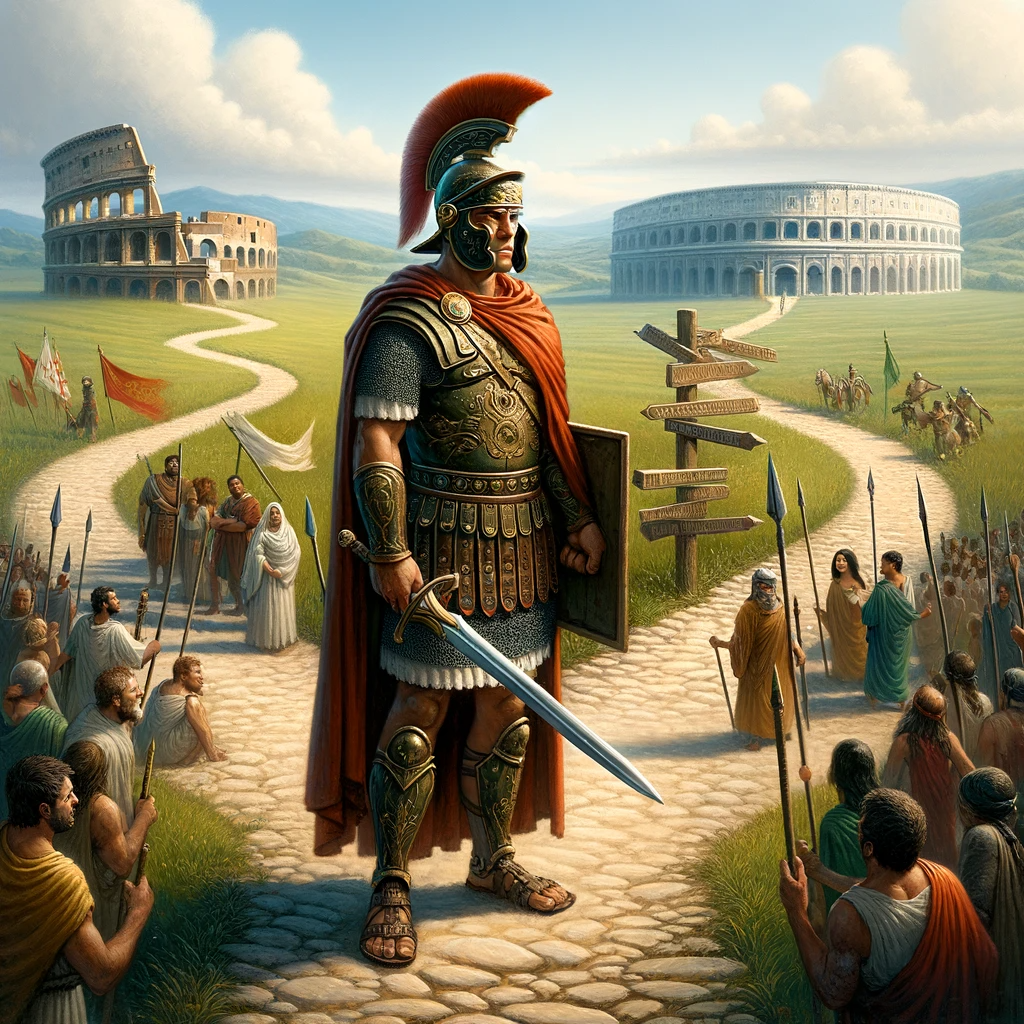}}
    \newline
    \vspace{0.2cm}
    
    \keyword{\textbf{Story B:}} In the heart of the Roman Empire, General Lucius found himself \redunline{torn between two irreconcilable paths}. On one hand, he had sworn an oath of unwavering loyalty to the empire that had granted him power and prestige. On the other, he witnessed the relentless oppression and suffering of the common people, burdened by heavy taxes and brutal tyranny. \redunline{The call for justice and equality} resonated within him like a thunderous battle drum. In a moment of agonizing decision, he chose to forsake his allegiance to the empire, forsaking his name and status to lead a rebellion for the oppressed, determined to wield his military prowess to carve a path toward a more just and equitable future.
    \\
    \midrule
    \rowcolor{shadecolor}
    \multicolumn{2}{>{\arraybackslash}p{\dimexpr\linewidth-2\tabcolsep}}{
    \keyword{\textbf{Review by GPT4:}}
    Story $A$ is better than $B$.
    
    Story $A$ is written with \greenunline{a more vivid and immersive narrative style}.
    It provides greater detail about \greenunline{the protagonist's character, motivations, and the historical context}. 
    Phrases like "vowed to use his immortality to dismantle the corrupt regime" and "a beacon of righteous rebellion against tyranny" \greenunline{add a mythical and heroic dimension to the protagonist, Lucius}.
    The mention of a vengeful deity granting immortality to Lucius \greenunline{adds a layer of mythological intrigue and depth to his character}.
    Furthermore, the depiction of the empire's corruption and Lucius's transition from loyalty to rebellion is \greenunline{more dramatically and emotionally engaging, enhancing the reader's investment in the story}.
    
    Story $B$, while also well-written, \redunline{lacks the same level of detail} and \redunline{narrative flair}. It presents \redunline{a more straightforward account} of Lucius's dilemma and decision. The story \redunline{misses opportunities to deepen the character development and historical context}, resulting in a narrative that \redunline{feels less engaging and impactful} compared to $A$.
    Additionally, $B$ \redunline{repeats certain ideas}, such as Lucius being torn between loyalty and justice, which can \redunline{make the plot feel slightly redundant}.
    } \\
    \bottomrule
    \end{tabularx}
    \caption{
    This example highlights the significance of story premises.
    $A$ is a premise created by MoPS, and $B$ is its manually simplified form.
    These premises guide \texttt{Dall-E} and \texttt{chatgpt} in generating posters and stories, with \texttt{gpt-4-turbo} reviews emphasizing the role of a well-crafted premise in developing engaging narratives.}
    \vspace{-0.5cm}
    \label{tab:motivation}
\end{table*}
Premise is what your story is about~\cite{field2005screenplay, lyons2015anatomy}.
A story premise is a concise line
that captures the story's main idea, conflict, and characters, outlining its foundation and direction ~\cite{truby2008anatomy,cron2012wired,brody2018save}.
Writers use the premise to guide story development, offering strategic insight into characters, plot, theme, and resolution.
In Automatic Story Generation (ASG), substantial research has explored various systems
~\cite{DBLP:conf/acl/LewisDF18, DBLP:conf/acl/FanLD19,DBLP:conf/aaai/YaoPWK0Y19, DBLP:conf/emnlp/YangTPK22, DBLP:conf/acl/YangKPT23, DBLP:journals/corr/abs-2305-13304}.
These systems need input to trigger and guide story creation.
A premise serves as such an input, offering a starting point for complex narrative development.
However, crafting a story premise challenges artistic and technical skills, requiring the capture of core elements and appeal in minimalistic text~\cite{lyons2015anatomy}.
A strong dramatic premise is fundamental to most successful stories~\cite{truby2008anatomy}.
In Tab.~\ref{tab:motivation}, we illustrate the significance of a fascinating story premise in creating engaging narratives.
If we can automate the design and creation of diverse and high-quality premises, it would be a major boost to the field of story generation.
Most future ASG frameworks could benefit from using these generated premises to thoroughly and comprehensively evaluate the effectiveness of their frameworks.

Existing work primarily obtains story premises through the following three methods:
(1) \textbf{Dataset Premise Extraction}: randomly extracting ready-made story premises from public datasets~\cite{DBLP:conf/acl/FanLD19, DBLP:conf/aaai/YaoPWK0Y19, DBLP:conf/naacl/TanYAXH21}, such as WritingPrompts (WP)~\cite{DBLP:conf/acl/LewisDF18}.
However, it suffers from inconsistent quality, including nonsensical premises, and offers limited customization.
(2) \textbf{LLM Premise Induction}: utilize models' extensive knowledge to generate numerous story premises~\cite{DBLP:conf/emnlp/YangTPK22,DBLP:conf/acl/YangKPT23, DBLP:journals/corr/abs-2310-08796}.
Its drawback lies in an over-reliance on the model's knowledge base, potentially curtailing the diversity and innovation of the generated premises~\cite{DBLP:journals/corr/abs-2309-05196}.
(3) \textbf{Human-Curated Premise}: depend on premises provided or predefined by humans~\cite{rosa-etal-2022-gpt,DBLP:conf/chi/MirowskiMPE23}.
The significant flaw here is the time-consuming and labor-intensive nature of manually writing premises, especially when generating stories in bulk.
Overall, current research area lacks a reliable automated method for generating premises.
In this paper, we still adopt the approach of inducing from LLMs with extensive world knowledge via prompts.
However, we focus on inducing fine-grained modules.
Our novelty lies in creative combinations of modules to generate a large number of diverse, fascinating, complete, and original story premises.
Based on this, we introduce \textbf{Mo}dular Story \textbf{P}remise \textbf{S}ynthesis (MoPS). 
It deconstructs a complete premise into modules, gathers module candidates into a hierarchical structure, outlines a premise design from selected elements, and finally has LLM synthesize these into a cohesive story premise sentence (\S\ref{sec:method}).
Our evaluations (\S\ref{subsec: evaluation_on_premises}) show that premises we've created stand out on various quality and diversity criterion (\S\ref{subsec:criteria}), surpassing those generated by LLMs or sourced from public story datasets.
Generated premises, when integrated with state-of-the-art story generation pipelines~\cite{DBLP:conf/chi/MirowskiMPE23, DBLP:journals/corr/abs-2305-13304}, not only yield tailored narratives but enhance the overall quality of resulting stories (\S\ref{subsec:evaluation_on_story}).

This paper pioneers the modular synthesis of story premises.
Our work aims to contribute to the field of ASG in the following ways:

(1) Highlighting the critical role of premises in story generation, and encouraging a deeper focus on the design and creation of story premises.

(2) Introducing MoPS, a method for automated design and creation of premises, along with two metrics for premise diversity and three for quality, conducting a thorough evaluation of our premises.

(3) Grafting two story generation pipelines for our premises and offering three version datasets: curated (100 premise-story pairs), moderate (1k premise-story pairs), and complete (7.6k premises).

\section{Related Work}
\label{sec:related_work}

\subsection{Automatic and Controllable Story Generation via Premise}
\label{subsec:automatic_story_generation}
\textbf{Dataset Premise Extraction.} \citet{peng-etal-2018-towards} and \citet{DBLP:conf/aaai/YaoPWK0Y19} used a word from each ROCStories (ROC) sentence as a premise to generate short stories.
\citet{DBLP:conf/acl/LewisDF18} defined premises as topic-describing sentences, conditioning story generation on them.
They also created the WP dataset with 300k premise-story pairs from Reddit.
\citet{DBLP:conf/acl/FanLD19} used WP premises as inputs, adding a predicate-argument structure for enhanced coherence.
Furthermore, many studies~\cite{DBLP:conf/emnlp/XuPSPFAC20,DBLP:conf/naacl/TanYAXH21,DBLP:conf/icml/PapalampidiCK22,DBLP:conf/naacl/HanCTP22,DBLP:conf/coling/SunSMLF22,chen-etal-2022-coherent,DBLP:conf/uist/Peng0HZMQ23,li2023enhancing,DBLP:journals/corr/abs-2310-08185,DBLP:conf/emnlp/HuangQLWBCC23,DBLP:conf/emnlp/WangYLK23} use ROC or WP premises as initial triggers in story generation.
Public dataset premises vary in quality without a unified standard, with nonsensical premises, including nonsensical examples found in WP and ROC.
This variability can impact story quality, obscuring framework performance.
Our paper identifies essential premise elements and establishes synthesis standards to ensure their completeness.

\noindent\textbf{Human-Curated Premise.} Some works employ manually provided story premises~\cite{rosa-etal-2022-gpt,DBLP:conf/chi/MirowskiMPE23, DBLP:journals/corr/abs-2305-13304,DBLP:journals/corr/abs-2310-12902}.
For example, \citet{DBLP:conf/chi/MirowskiMPE23} employs loglines for hierarchical script generation.
\citet{DBLP:journals/corr/abs-2310-08185} uses genres and themes as premises for rolling generated novels.
Manual premise selection is limited in number and scalability, may leading to bias.
MoPS generated up to 7.5k premises cost-effectively.
We validated premises' importance for LLM-based generation by using Dramaton~\cite{DBLP:conf/chi/MirowskiMPE23} and RecurrentGPT~\cite{DBLP:journals/corr/abs-2305-13304} to produce scripts and novels.
Based on this, we created and publicly released datasets containing pairs of premises and corresponding stories.

\noindent\textbf{LLM Premise Induction.}
Currently, inducing premises from LLMs via prompts (e.g., ``Write a premise for a short story.'') is mainstream.
Recent works increasingly use LLM-written premises, leveraging LLMs' extensive knowledge~\cite{DBLP:conf/emnlp/YangTPK22, DBLP:conf/acl/YangKPT23, DBLP:journals/corr/abs-2310-03304, DBLP:journals/corr/abs-2310-08796}.
Despite their language capabilities, LLMs face criticism for potentially less diverse and repetitive contents~\cite{DBLP:journals/corr/abs-2309-05196,  DBLP:journals/corr/abs-2309-14556, meincke2024prompting}.
MoPS narrows focus by inducing specific modules (e.g., persona, main events) from LLMs, unlike direct premise induction.
This approach enables  creators to creatively combine candidates from modules, producing unique and innovative outputs.

\subsection{Textual Data Synthesis via Large Language Models}
\label{subsec:textual_data_synthesis}

Synthesizing textual data with off-the-shelf LLMs is a new trend in data engineering~\cite{DBLP:conf/acl/WangKMLSKH23,DBLP:journals/corr/abs-2304-12244}.
Synthesized data shows promise in model training, reducing hallucinations, and enhancing mathematical reasoning.
\citet{DBLP:journals/corr/abs-2305-07759} used specific verbs, nouns, and adjectives to have \texttt{gpt-3.5-turbo} generate short stories for 3-4 year-olds.
\citet{DBLP:journals/corr/abs-2306-11644} synthesized Python textbooks by defining their theme and target audience.
This synthesis method was also applied to common sense reasoning data~\cite{DBLP:journals/corr/abs-2309-05463}.
\citet{DBLP:journals/corr/abs-2312-09241} used \texttt{gpt-3.5-turbo} to expand \texttt{GSM8K} dataset questions into more math word problems.
\citet{DBLP:conf/emnlp/RadharapuRAL23} synthesized safety test data for LLMs using harmful task categories, policy concepts, and geographic regions.
Our work uniquely focuses on synthesizing story premises.
MoPS specifies meaningful modules within story premises.
Crucially, our modules have sequential dependencies, like persona depending on the background and theme.
This interlinks modules into a nested dictionary.
We demonstrate (\S\ref{subsec: ablation}) that sequential dependencies are vital for consistent story premises in ablation experiment.

\section{Modular Story Premise Synthesis}
\label{sec:method}

\begin{figure}[t] \centering
    \includegraphics[width=\linewidth]{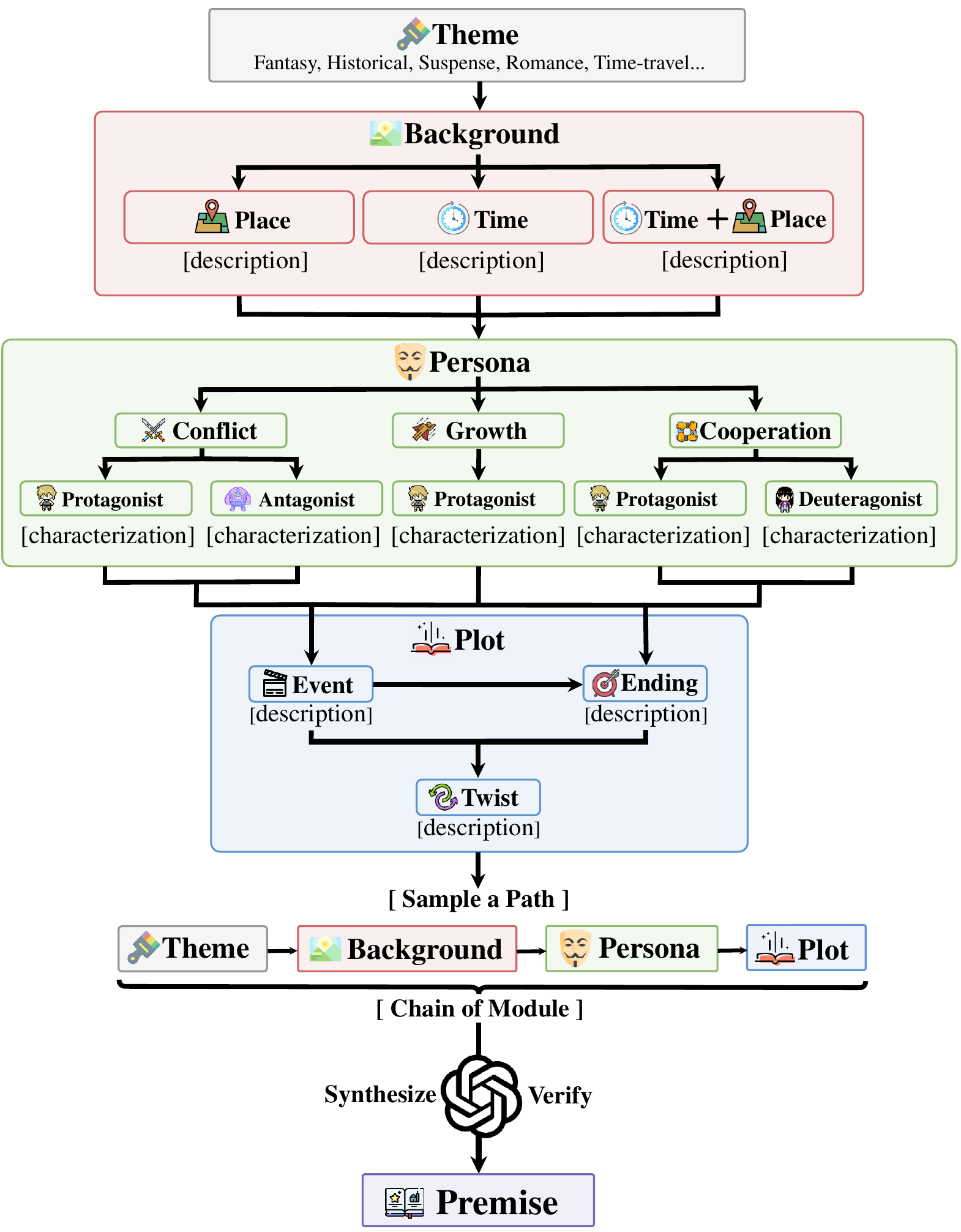}
    \caption{Overview of MoPS. We divide the premise into four ordered modules: \theme{theme}, \background{background}, \persona{persona}, and \plot{plot}, with each module further divided into submodules. 
    From the top down, arrows indicate the dependency relationships within and between modules.} \label{fig:mops}
\end{figure}

\subsection{Overview}

Fig.~\ref{fig:mops} illustrates the overview of MoPS, which dissects a premise into sequentially dependent modules.
Its core idea is to transform the design of premise into sampling from candidates within each module, converting open-ended generation from scratch into the synthesis of modular elements.

\textbf{Anatomy of Story Premise.} A premise, which outlines what a story is about, should contain elements similar to those in a story.
We divide the premise into four ordered modules: \theme{theme}, \background{background}, \persona{persona}, and \plot{plot}, with each module further divided into submodules.
This entails subdividing background into time, place, and their combination, persona into three categories: growth, conflict and collaboration, plot into event, ending, and twist.

\textbf{Dependency between Modules.}
The arrows in Fig.~\ref{fig:mops} illustrate the dependency between and within modules, following the natural logic of story construction.
Initially, determining the theme of a premise ensures that all following modules serve a unified central idea.
Next, background setting provides temporal and spatial context for the premise, offering a stage for persona and plot modules.
Persona is the core of premise, driving plot forward through characters' behaviors and decisions.
Plot is the main body of story development, with a main event forming the backbone that runs through the narrative, a ending provides a clear resolution and ensures a closed loop of premise, a twist that can enhance premise's appeal and makes it engaging.

\textbf{Insight behind Modular Design.} The effectiveness of MoPS primarily stems from its modular design, embodying the concept of \textit{combinatorial creativity}~\cite{DBLP:conf/semweb/SuchanekMBC16a,DBLP:conf/aaai/GuzdialR18,simonton2021scientific}. 
That is, while each component may represent existing ideas, their combination can boost unique and innovative outcomes.
MoPS's effectiveness is specifically manifested in its ability to produce diverse, fascinating, complete, and original story premises.

\subsection{Induce Candidates from LLM}
\label{subsec:induce}
We instruct LLM to act as a creator, generating candidates for each module.
Since ingredients are not independent but sequentially dependent, we reflect this dependency via prompts.
The induction prompt for each component will incorporate a candidate from each preceding module as a precondition.
For example, when collecting event candidates, the prompt (see Tab.~\ref{tab:prompt_for_inducing_events}) will include a theme, a background, and a persona, thereby instructing LLM to generate plausible events and descriptions under these preconditions.

Formally, we first manually pre-define a group of theme candidates $\mathcal{C}_{\alpha}=\{\alpha_1, \alpha_2,...,\alpha_m\}$.
For \textbf{each} theme $\alpha_i$, we collect compatible background candidates $\mathcal{C}_{\beta | \alpha_i}=\{\beta_{1 | \alpha_i}, \beta_{2 | \alpha_i},...,\beta_{n | \alpha_i}\}$ that may appear under that $\alpha_i$.
Likewise, we gather compatible persona candidates $\mathcal{C}_{\gamma | \beta_j, \alpha_i}$ for each $\beta_j$ and $\alpha_i$.
Similarly, we can obtain event candidates $\mathcal{C}_{\delta | \gamma_k, \beta_j, \alpha_i}$, ending candidates $\mathcal{C}_{\omega | \delta_l, \gamma_k, \beta_j, \alpha_i}$ and twist candidates $\mathcal{C}_{\sigma | \omega_t, \delta_l, \gamma_k, \beta_j, \alpha_i}$.

\textbf{Data Structure of Module Candidates.}
The induction process essentially forms a nested dictionary $\mathcal{D}$.
The first layer is the theme dictionary, where each key is a theme candidate, and each value is the corresponding background dictionary for that theme.
Subsequently, persona, event, ending, and twist dictionaries are nested in sequence.
Sampling a key path from  $\mathcal{D}$ serves as the \textit{design of premise}.
By performing a pre-order traversal of the entire nested dictionary, we can achieve a wide variety of combinations of module candidates, significantly fostering combinatorial creativity to generate unique and innovative story premises.

\textbf{Deduplication for Module Candidates.} In light of recent concerns over repetitiveness of LLM creativity~\cite{DBLP:journals/corr/abs-2309-05196,  DBLP:journals/corr/abs-2309-14556, meincke2024prompting}, we employ embedding similarity~\cite{DBLP:conf/emnlp/ReimersG19} for deduplication whenever a new candidate joins.
For pairs of candidates with a cosine similarity greater than threshold $\epsilon$, we retain only one.

\textbf{Resilience for Human-in-the-Loop}. It's worth noting that this process is not exclusive to LLMs. Human creators can follow the same method, sequentially coming up with each component and then synthesizing a premise with the aid of the linguistic capabilities of language models.

\subsection{Synthesize and Verify Story Premise}
\label{subsec:synthesize_and_verify}

\begin{figure}[t] \centering
    \includegraphics[width=\linewidth]{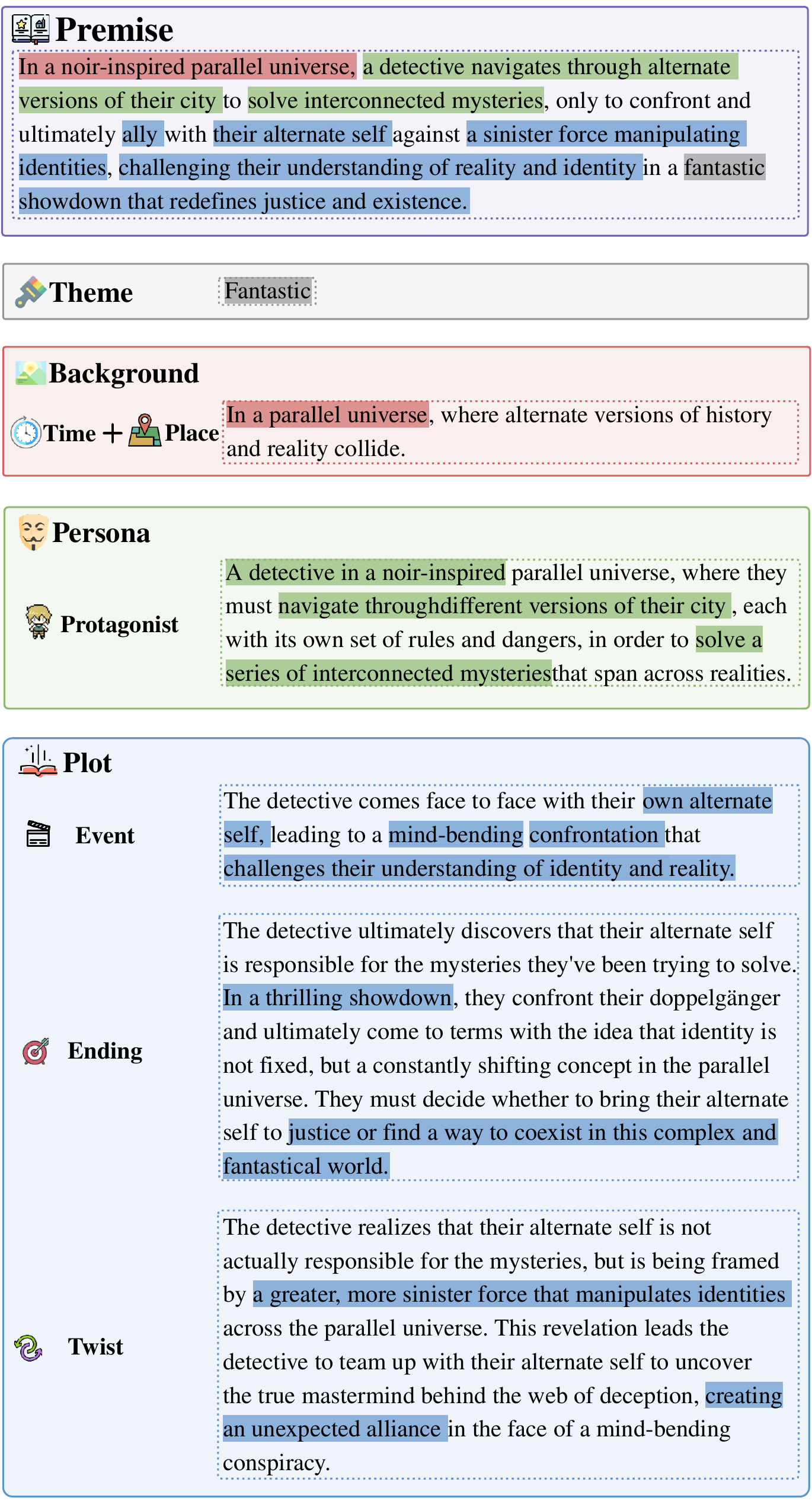}
    \caption{Case study on premise synthesis demonstrates LLM's ability to extract core information from modules and integrate them into a cohesive final premise, effectively encapsulating the sampled module path.}
    \label{fig:synthesis_case}
\end{figure}

In \S\ref{subsec:induce}, we construct a nested dictionary holding candidates for each module.

\textbf{Premise Synthesis.}
Sample a key path from the nested dictionary as the design of premise, we instruct LLM to meld the design of premise into a compact, concise and coherent sentence as the story premise.
The synthesis prompt is shown in Tab.~\ref{tab:prompt_for_synthesis}. 
Additionally, we provide a case study of premise synthesis in Fig.~\ref{fig:synthesis_case}.
Notably, during the synthesis process, certain modules can be flexibly excluded by adding a mask. This means setting the candidate of the corresponding module in the key path to an empty string.

\textbf{Self-Verification.}
As mentioned above, we incorporate a 
candidate from each preceding module as preconditions in the prompt to (softly) constrain LLM's output.
However, due to the inherent hallucination of LLM~\cite{DBLP:journals/corr/abs-2309-01219,DBLP:journals/csur/JiLFYSXIBMF23}, there is still a small chance of generating candidates that are inconsistent with preconditions or contain factual errors.
This may diminish the quality of the synthesized premises.
Therefore, following the synthesis procedure, we further instruct LLM to self-verify whether synthesized premises contain any obvious inconsistencies or factual errors (see prompt in Tab.~\ref{tab:prompt_for_verification}). 
If so, that corrupt premise will be discarded.

Integrating \S\ref{subsec:induce} and \S\ref{subsec:synthesize_and_verify}, MoPS first pre-collects a vast number of candidates for each module, forming a nested dictionary.
Then, a key path (theme, background, persona, event, ending, and twist) is sampled as the design for premise.
Finally, LLM is driven to synthesize items in the path into a fluid sentence serving as the story premise.

\section{Experiment Settings}
\label{sec:exp_settings}

\subsection{Dataset Construction}
\label{subsec:dataset}
We derive module candidates from \texttt{gpt-3.5-turbo}.
Initially, we collect 14 narrative themes from well-known novel and drama websites.
For each theme, we gather 30 background candidates, 10 for each time, place, and both.
For each background, we collect 9 personas, 3 for growth, conflict, and cooperation each.
We then prepare 2 main events for each persona.
For each event, we construct a final ending.
Finally, for each event-ending pair, we conceive a twist.

\noindent\textbf{Complete Version}.
The previous step produces a nested dictionary.
By performing a pre-order traversal, we obtain a total of 7,600 premise designs.
These designs are synthesized into premises and then verified by \texttt{gpt-3.5-turbo}.
We get 7,599 valid story premises, showing that injecting preceding premise modules into prompts can largely prevent inconsistencies and factual errors.
All these story premises constitute the complete version.

\noindent\textbf{Moderate Version}.
We randomly select 1,000 entries from the complete version to validate MoPS's ability to synthesize diverse and high-quality story premises.
Evaluation metrics are detailed in \S\ref{subsec:criteria}.
We integrate two advanced story generation frameworks, Dramatron and RecurrentGPT, for generating scripts and novels, with \texttt{gpt-3.5-turbo} serves as the language backend.
The 1,000 premise-story pairs, each including a novel and a script, comprise moderate version.
Dramatron parameters follow those in its original paper.
RecurrentGPT's iteration number is set to 10.
The scripts averaged about 5k tokens, and novels 2.2k tokens.

\noindent\textbf{Curated Version}.
From moderate version, we select a diverse, high-quality subset.
It includes 100 premise-story pairs.
Selection details are in \S\ref{appendix:curated_dataset}.
Synthesized premises can serve as a benchmark for evaluating subsequent story generation methods.
Generated novels and scripts are useful for pre-training or fine-tuning language models, especially junior models~\cite{DBLP:journals/corr/abs-2305-07759}, enhancing storytelling end-to-end~\cite{DBLP:journals/corr/abs-2310-08796}.

\subsection{Criteria for Premise Diversity and Quality}
\label{subsec:criteria}

\begin{figure}[tbp]
\centering
\begin{subfigure}[htb]{0.48\linewidth}
  \centering
  \label{fig:mops_breadth_score}
  \includegraphics[width=1.09\linewidth,]{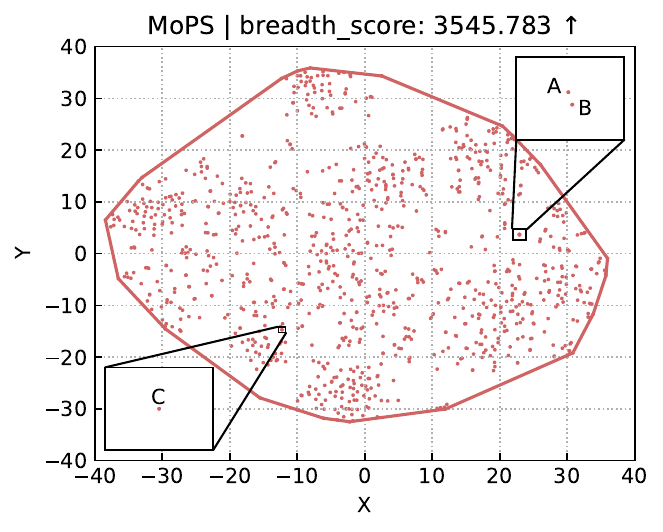}
  \vspace{-20pt}
\end{subfigure}
\hspace{1pt}
\begin{subfigure}[htb]{0.48\linewidth}
  \centering
  \label{fig:mops_density_score}
  \includegraphics[width=\linewidth]{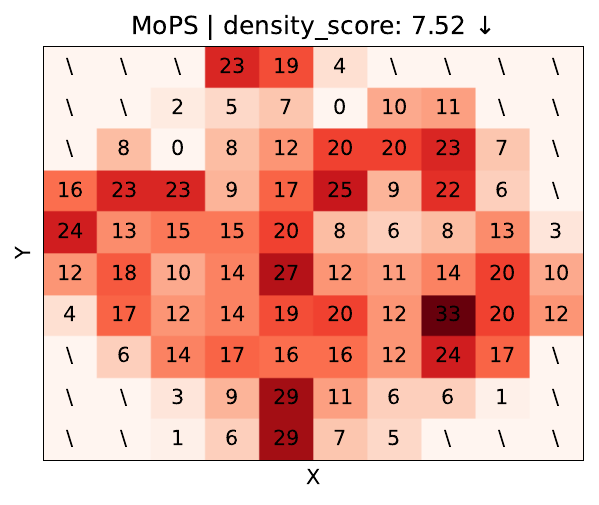}
  \vspace{-15pt}
\end{subfigure}

\begin{subfigure}[htb]{0.99\linewidth}
  \centering
  \label{fig:diverse_case_study}
  \includegraphics[width=\linewidth]{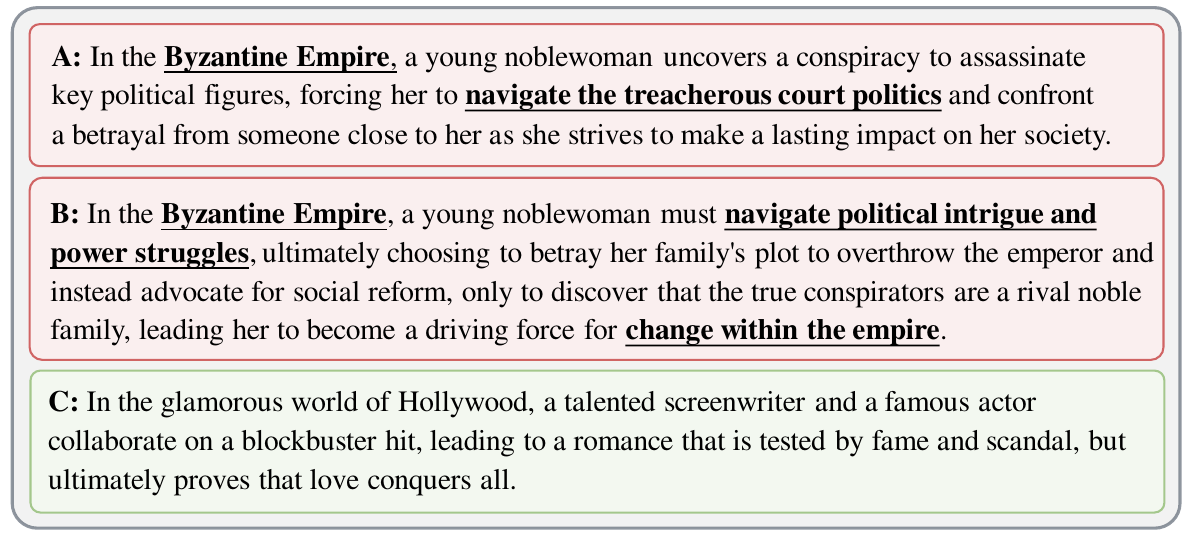}
\end{subfigure}
\caption{Diversity Metrics.
Breadth score, shown top left, measures the polygon area from 2D semantic embedding vectors. Density score, displayed top right, calculates the standard deviation \textbf{within} the polygon from a 2D histogram. Examples (A, B, C) illustrate that reduced-dimension embeddings effectively capture semantic similarity.}
\label{fig:diversity_metrics}
\end{figure}

To effectively assess synthesized story premises as open-ended texts with defined semantics, we introduce five automated evaluation metrics: two for diversity and three for quality.

\noindent\textbf{Diversity Metrics}.
As shown in Fig.~\ref{fig:diversity_metrics}, we focus on the breadth and density of semantic coverage.
To quantify these indicators, we utilize \textit{all-MiniLM-L6-v2} from sentence-transformers~\cite{DBLP:conf/emnlp/ReimersG19} to encode premises into high-dimensional embedding vectors and employ t-SNE~\cite{van2008visualizing} to reduce them to a 2D plane.
\textbf{Breadth} is defined as the area of embedding polygon:
\begin{align}
    \label{eq:breadth}
    \text{Breadth Score} &= f(\{(x_i, y_i) \mid i \in [0, N)\}) \\
    (x_i, y_i) &\in \text{t-SNE}(\text{Embedding}(\text{Premises})) \notag 
\end{align}
where $f$ represents a function for calculating the area of a polygon formed by $N$ semantic vectors $(x_i, y_i)$, implemented by the shapely~\cite{GilliesShapely2023} library.
Area is larger for better.
\textbf{Density} is defined as the standard deviation of the count sequence in the 2D histogram of embedding polygon:
\begin{align}
    \label{eq:density} 
    \text{Density Score} &= \sigma\left(\{c_{ij} ~|~i,j \in [0, M) \}\right) \\
    c_{ij} &= \text{Count}(b_{ij})~\text{where}~b_{ij} \in B \notag
\end{align}
where  $M$ is the number of bins per row and column, set to 10.
$\sigma$ is the standard deviation.
$B$ denotes the set of bins within embedding polygon, and $c_{i,j}$ is the count for bin $b_{ij}$.
A lower value indicates the number of points in each bin is similar, reflecting a higher uniformity of coverage, and vice versa.

In \S\ref{sec:further_study_diversity_metrics}, we conduct further study on the reliability of diversity metrics. The results show that: 1) The evaluation scores are relatively robust across different t-SNE random states and sentence-transformer models, and 2) The diversity evaluation results generally align with human intuition.

\noindent\textbf{Quality Metrics}.
Referencing numerous previous works~\cite{DBLP:journals/corr/abs-2302-04166, DBLP:journals/corr/abs-2305-13304, DBLP:journals/corr/abs-2310-08185, DBLP:journals/corr/abs-2310-00785}, we employ  \texttt{gpt-4-turbo} with temperature=0 as a versatile scorer, and tailor three dimensions for story premise:
(1) \textbf{Fascination}, measuring whether the premise is sufficiently interesting;
(2) \textbf{Completeness}, assessing whether the premise possesses its requisite components;
(3) \textbf{Originality}, gauging the level of familiarity to LLM, with the expectation that story premise is unfamiliar to language models.
The scoring range is $[0,100]$, with higher scores being better.
All prompts are listed in \S\ref{appendix: quality_prompts}.

\subsection{Baselines}
\label{subsec:baselines}
We select 5 baselines to benchmark the superiority of our synthesized premises in terms of quality and diversity.
The first two are based on LLM, and the last three are based on public datasets.

\noindent\textbf{Vanilla (VIL)}: instruct \texttt{gpt-3.5-turbo} (temperature=0.6) to create premises and deduplication.

\noindent\textbf{Complex (CPX)}: similar to VIL, but with 3 MoPS premises as few-shot examples.

\noindent\textbf{DOC}: from~\citet{DBLP:journals/corr/abs-2310-08796}, generated by prompting to \texttt{llama2-13b-chat}.

\noindent\textbf{WritingPrompts (WP)}: collected by~\citet{DBLP:conf/acl/LewisDF18} from Reddit, with premises being real user-written story post titles.

\noindent\textbf{Storium (STM)}: collected by~\citet{DBLP:conf/emnlp/AkouryWWHPI20} from a HCI card game platform, we use the description of the storytelling game as the premise.

For MoPS, we use \textbf{moderate} version, which contains 1,000 premises.
For LLM-based baselines, we induce and deduplicate  until reaching 1,000 premises.
For public datasets, we randomly extract 1,000 premises for evaluation.
We provide more details about each baseline in \S\ref{appendix: baselines}.

\begin{table*}[htbp]
\centering
\resizebox{\textwidth}{!}{
\begin{tabular}{@{}lccccccc@{}}
\toprule
\multirow{2}{*}{} & \multirow{2}{*}{Tokens} & \multicolumn{2}{c}{Diversity Score} & \multicolumn{4}{c}{Quality Score (mean $\pm$ std)} \\ \cmidrule(l){3-4} \cmidrule(l){5-8} 
                     &                         & Breadth$\uparrow$          & Density$\downarrow$           & Fascination$\uparrow$ & Completeness$\uparrow$ & Originality$\uparrow$ & Average$\uparrow$ \\ \midrule
MoPS  &  50.24 & \textbf{3545.78} & \textbf{7.52} & \textbf{75.66} $\pm$ 3.98 & \textbf{74.78} $\pm$ 3.16 & \underline{60.01} $\pm$ 23.61 & \textbf{70.15} $\pm$ 10.25
 \\
CPX  & 45.22 & 2618.18 &  14.63 &  73.96 $\pm$ 3.95 & \underline{70.88} $\pm$ 6.33 & 26.60 $\pm$ 12.94 & 57.15 $\pm$ 7.74~~ \\
VIL  & 37.61 & \underline{3050.72} & 11.08 &  71.50 $\pm$ 5.95  & ~~53.66 $\pm$ 13.20 
 &  20.82 $\pm$ 11.71 & 48.66 $\pm$ 10.29\\
DOC  & 45.81 & 1901.04 & 14.22 & \underline{74.33} $\pm$ 6.31 &  ~~69.87 $\pm$ 11.80 &  50.65 $\pm$ 27.57 &  64.95 $\pm$ 15.23\\
WP  & 42.54  & 3013.61 & \underline{8.53} & ~~\underline{74.49} $\pm$ 13.65 &  ~~43.57 $\pm$ 17.46  & \textbf{71.05} $\pm$ 30.18 & 63.04 $\pm$ 20.43\\
STM  & 77.78 & 1918.67 & 13.82 & ~~\textbf{75.83} $\pm$ 10.79 &  ~~66.45 $\pm$ 16.30 &  \underline{61.51} $\pm$ 30.62 & \underline{67.93} $\pm$ 19.24\\
\bottomrule
\end{tabular}}
\caption{Diversity score and quality score of story premises.
The plots for diversity scores are in Fig.~\ref{fig:breadth_score},~\ref{fig:density_score}, for quality scores are in Fig.~\ref{fig:fascination_score},~\ref{fig:completeness_score},~\ref{fig:originality_score}.
The optimal values (with $p$ < 0.05) will be \textbf{bolded}, and the second-best values (with $p$ < 0.05) will be \underline{underlined}.
}
\label{tab:premise_scores}
\vspace{-5pt}
\end{table*}

\section{Experiment Results and Analysis}
\label{sec:exp_results}
Our experiments focus on three scenarios:
(1) \textbf{Evaluate premise diversity and quality:} Whether MoPS produces more diverse and higher-quality premises than baselines.
(2) \textbf{Component ablation:} The effectiveness of MoPS's modular design and sequential module dependencies.
(3) \textbf{Long story quality assessment:} Whether MoPS premises yield higher quality long stories.

\textbf{Significance Tests.}
We conduct significance tests to verify results' validity.
Our significance tests evaluate:
(a) if MoPS's premises and long stories outperform baselines;
(b) if masking MoPS components impacts premise quality.

\subsection{I: Evaluation on Story Premises}
\label{subsec: evaluation_on_premises}
As introduced in \S\ref{subsec:baselines}, we evaluate 1,000 premises from MoPS and baselines.
The results are shown in Tab.~\ref{tab:premise_scores}.
For diversity, semantic breadth and density are calculated (Eq.~\ref{eq:breadth}, \ref{eq:density}).
Illustrative diversity score diagrams in \S\ref{appendix:diversity_fig} due to space limits.
The plot of quality scores assessed by \texttt{gpt-4-turbo} (see prompts in \S\ref{appendix: quality_prompts}) can be found in \S\ref{appendix:quality_fig}.
Mean and standard deviation for each dimension are reported, with significance testing on means.
Reporting standard deviation explores the evenness of premise quality.
The main observations are listed as follows:

(1) \textbf{The introduction of MoPS's modular design significantly enhances semantic diversity.}
MoPS consistently outperforms all baselines in semantic breadth and density (Tab.~\ref{tab:premise_scores}, Fig.~\ref{fig:breadth_score}).
\textbf{Breadth}: MoPS's semantic polygon area is 1.865x larger than DOC and 1.162x larger than VIL, with an average of 1.481x.
\textbf{Density}: MoPS scores are 48.6\% better than CPX and 11.8\% better than WP, averaging a 37.1\% improvement.
Lower density scores indicate more uniform distribution of semantic vectors, reflecting better diversity.
The diversity gain stems from the modular design, enabling deeper focus on innovation and depth of each part, overcoming the complexity of creating an entire story premise.
Traversing the pre-collected nested dictionary achieves diverse premise designs, as reflected in the diversity scores.

(2) \textbf{MoPS can generate fascinating, complete, and original story premises.}
MoPS outperforms most baselines in fascination, completeness, and originality (Tab.~\ref{tab:premise_scores}, Figs.~\ref{fig:fascination_score}-\ref{fig:originality_score}).
(a) \textbf{Fascination}: MoPS and STM score similarly ($p$=0.64) and surpass other baselines. Unlike STM's reliance on human-in-the-loop, MoPS's premise requires just one API call or model inference.
(b) \textbf{Completeness}: MoPS, by specifying premise ingredients, scores higher than all baselines.
Notably, we left the definition of a complete premise to GPT-4's discretion without suggesting components (see Tab.~\ref{tab:prompt_for_completeness}).
Results affirm the premise design aligns with GPT-4's understanding, validating the modules' rationality.
The ablation in \S\ref{subsec: ablation} can further validate this argument.
(c) \textbf{Originality}: Assessed by querying GPT-4 on premise familiarity (see Tab.~\ref{tab:prompt_for_originality}), indicating uniqueness.
Results show that MoPS competes originality with human-written premises (WP, Storium).
VIL and CPX premises, generated by \texttt{gpt-3.5-turbo}, are familiar to \texttt{gpt-4-turbo}, resulting in low scores.
Despite ingredients of MoPS premises also come from \texttt{gpt-3.5-turbo}, their combination significantly enhances originality (2-3x compared to VIL and CPX).
These outcomes confirm that modular design and creative module combinations yield unique, innovative outputs.

(3) \textbf{The quality of the premises is more homogeneous.}
MoPS shows lower standard deviations across three dimensions than most baselines (Tab.~\ref{tab:premise_scores}).
In contrast, human-written premises (WP and STM) exhibit significant fluctuations.
This consistency is due to MoPS's modular design specifying components, akin to assembly line products.

In addition, we also conducted human and \texttt{claude-3-opus} evaluation. The results and analysis are detailed in \S\ref{sec:more_evaluation}. These findings are consistent with those evaluated by \texttt{gpt-4-turbo}, affirming the reliability of powerful LLM evaluation.

\subsection{II: Ablation on Modules and Dependence}
\label{subsec: ablation}

\begin{table}[t]
\centering
\small
\resizebox{\columnwidth}{!}{
\begin{tabular}{@{}lccccc@{}}
\toprule
& Fascination$\uparrow$ & Completeness$\uparrow$ & Originality$\uparrow$ \\ \midrule
MoPS                      &  \textbf{75.81}           &  \textbf{75.10}            & \underline{59.90}           \\  \midrule
m/f Twist                      &  \underline{74.56}           &  \underline{73.20}            &  41.90           \\
m/f Ending                      & 74.43           & 71.40            &  42.70           \\
m/f Event                      &  74.16           &  67.20            &  39.10           \\ 
m/f Persona                      & 73.30           &  57.25            &   26.90           \\  \midrule
w/o Dependence                      &  65.77          & 65.75            &  \textbf{74.51}           \\
\bottomrule
\end{tabular}}
\caption{Ablation results. m/f $\rightarrow$ ``mask all components following a certain component''. w/o $\rightarrow$ ``without''.}
\label{tab:ablation_scores}
\end{table}

From the moderate dataset, we sample 100 premises and their designs.
Specifically, we aim to verify: (a) \textbf{Component effectiveness in premise design}.
We mask all components following a certain component (denoted as \textbf{m/f}) and re-synthesize the premise.
(b) \textbf{Dependency necessity between modules.}
We disrupt dependencies by cross-selecting components from all designs.
Results in Tab.~\ref{tab:ablation_scores}.
The main observations are as follows:
(1) Quality scores decrease with fewer components, showing each's importance in MoPS.
(2) Premises from designs lacking sequential dependencies show decreased fascination and completeness but increased originality.
This is because the inconsistency of design elements led to unique but subpar premises.

\subsection{III: Evaluation on Premise-based Stories}
\label{subsec:evaluation_on_story}

\begin{table}[t]
    \centering
    \resizebox{\columnwidth}{!}{
    \begin{tabular}{@{}lccc@{}}
    \toprule
    & Fascination$\uparrow$ & Completeness$\uparrow$ & Originality$\uparrow$ \\ \midrule
    MoPS-RecurrentGPT   & \textbf{74.60} & \textbf{60.30} & \underline{69.45}\\
    CPX-RecurrentGPT  &  \textbf{74.20} & \underline{56.05} & 45.60\\
    VIL-RecurrentGPT  &  \textbf{74.40} & \textbf{57.30} &  48.00\\
    DOC-RecurrentGPT  &  \underline{73.30} &  \textbf{57.60} &  \underline{66.75}\\
    WP-RecurrentGPT  & \textbf{74.40}&  55.45 & \textbf{81.15}\\
    STM-RecurrentGPT & 73.00 &  54.95 &  \underline{64.20}\\ \midrule
    MoPS-Dramatron  & \textbf{70.59} & \textbf{74.50} & \textbf{94.20} \\
    CPX-Dramatron  &  \textbf{70.24} & \textbf{74.50} & \textbf{92.60} \\
    VIL-Dramatron  &  \underline{67.92} & \textbf{74.30} &  \underline{83.50} \\
    DOC-Dramatron  &  \textbf{70.35} &  \textbf{74.00} &  \textbf{91.35}\\
    WP-Dramatron  & 62.90 &  62.95 & \textbf{92.35} \\
    STM-Dramatron  & \underline{68.29} &  \underline{70.40} &  \underline{84.80}\\
    \bottomrule
    \end{tabular}}
    \caption{Quality score of premise-based stories. The optimal values (with $p$ < 0.05) will be \textbf{bolded}, and the second-best values (with $p$ < 0.05) will be \underline{underlined}.}
    \label{tab:story_score}
    \vspace{-8pt}
    \end{table}

\textbf{We aim to verify its consistency in automated story generation.}
To our knowledge, it is the first experiment to explore the impact of story premises on the story generation, which is conducted across story premises from up to 6 different sources.
We first randomly select 100 novels and scripts from the moderate dataset.
From 1,000 baseline-generated premises, we sample 100 to generate scripts and novels.
Examples of two genres are in \S\ref{appendix:examples_of_novels_and_scripts}.
Finally, \texttt{gpt-4-turbo} scores these stories, with prompts in \S\ref{subsec:prompts_in_story_evaluation}.

The results are shown in Tab.~\ref{tab:story_score}.
MoPS shows the best overall performance.	
Of 6 values for 3 metrics across 2 genres, 5 are \textbf{bolded}, 1 \underline{underlined}.
These improvements solely stem from changes to story premises.
This confirms that for automated story generation methods, the high quality of MoPS premises can similarly reflect in generated long stories.
Although challenging to quantify the diversity of long stories, MoPS premises can infuse story generation with a wider range of components, such as backgrounds and personas.
Our research aims to inspire subsequent researchers to recognize the critical role of premises in story generation and encourage further empirical studies.

\subsection{IV: Comparison with Reference Stories in Existing Dataset}
\label{sec:appendix_reference}
Some existing story datasets collect human-written short stories (usually less than a few hundred words) as references for premise-based stories generation.
Here, we aim to verify whether the stories expanded from MoPS premises can surpass those reference stories in quality.

Specially, we use 100 story premises from MoPS to instruct \texttt{gpt-3.5-turbo} to write short stories.
These stories match the typical lengths seen in the ROC Stories (ROC)~\cite{rocstories} and WritingPrompts (WP)~\cite{DBLP:conf/acl/LewisDF18} datasets, both of which are commonly used in research.
For ROC, we limit the stories to 5 sentences and 80 words. For WP, we cap them at 500 words, aligning with the average story length in these datasets.
Then, we employ \texttt{gpt-4-turbo} to review stories created from MoPS premises and reference stories in ROC and WP, evaluating them on their fascination, completeness, and originality on a scale from 0 to 100.
Tab.~\ref{tab:reference_score} presents the evaluation results of 100 stories.
\begin{table}[t]
    \centering
    \resizebox{\columnwidth}{!}{
    \begin{tabular}{@{}lccc@{}}
    \toprule
    & Fascination$\uparrow$ & Completeness$\uparrow$ & Originality$\uparrow$ \\ \midrule
    MoPS-ROC & \textbf{69.09} & \textbf{43.87} & \textbf{67.30} \\
    Reference-ROC & 25.87 & 15.76 & 61.83 \\
    \midrule
    MoPS-WP & \textbf{73.88} & \textbf{58.78} & 83.90 \\
    Reference-WP & 60.88 & 32.18 & \textbf{94.23} \\
    \bottomrule
    \end{tabular}}
    \caption{Quality score of MoPS premise-based stories and reference stories. The optimal values (with p < 0.05) are \textbf{bolded}.}
    \label{tab:reference_score}
    \vspace{-8pt}
    \end{table}
The main observations are as follows:

(1) \textbf{The results show that stories created from MoPS premises match reference stories in originality and outperform them in fascination and completeness.}
Considering evaluation results presented in Tab.~\ref{tab:story_score}, we have grounds to believe that not only do long stories (>2000 words) extended from MoPS premises surpass 5 baselines we compared, but short stories expanded from MoPS premises also exceed references in existing story datasets.

(2) \textbf{As stories get longer, their fascination, completeness, and originality scores tend to rise (both in MoPS and Reference)}. For example, MoPS score for completeness increase from about 43.87 for a short MoPS-ROC story ($\approx$80 words) to 58.78 for a medium-length MoPS-WP story ($\approx$500 words), and then to 60\textasciitilde75 for a longer MoPS-RecurrentGPT/Dramatron story (>2000 words).
This is an interesting yet reasonable discovery since longer stories tend to include more captivating elements. This finding not only validates the rationality of the metrics designed in our work but also suggests that future research should explore longer stories.

\section{Conclusion}
\label{sec:conclusion}

This paper presents MoPS, a modular approach that automates the design and creation of story premises.
Using MoPS, we synthesized a large number of diverse and high-quality premises, generating extended novels and scripts.
Thorough evaluation demonstrates the superiority of MoPS over multiple baselines.
Similarly, extended stories from our premises also exhibit higher quality.
Based on our premises and extended stories, we created three versions of premise-story dataset to accommodate research for varied research scales.
Future ASG frameworks can benefit from these premises for thorough effectiveness evaluation.
We believe our research will advance the field of automated story generation.
Looking to the future, we hope to explore the impact of premises on cross-modal story creation, such as story poster generation~\cite{Dalle}, graphic narratives~\cite{internlmxcomposer2}, and even video stories~\cite{sora}.

\section{Limitations}
\label{sec:limitations}

\textbf{Balance Module Candidates.}
Inducing ending and twist modules, LLM tends to yield positive outcomes.
Yet, tragic works like "Les Misérables" remain popular.
Future work will include manually adding tragic endings and twists to enhance premise diversity in MoPS.

\noindent\textbf{More evaluation mechanisms.}
Considering concerns about reliable assessment of crowdsourcing platforms on open-ended text generation ~\cite{DBLP:conf/emnlp/AkouryWWHPI20,DBLP:conf/emnlp/KarpinskaAI21}, this paper, following many previous works~\cite{DBLP:journals/corr/abs-2302-04166, DBLP:journals/corr/abs-2305-13304, DBLP:journals/corr/abs-2310-08185, DBLP:journals/corr/abs-2310-00785}, employs powerful large language models and human as evaluators to assess premises and stories generated based on those premises.
Future work may explore diverse evaluation methods, including personalized story evaluation~\cite{DBLP:journals/corr/abs-2310-03304}, consulting with literary experts~\cite{DBLP:conf/chi/MirowskiMPE23}.

\section*{Acknowledgement}
We are grateful to anonymous reviewers for reviewing our paper and providing valuable feedback. We thank Zengzhi Wang, Ethan Chern, Xuefeng Li, Haoyang Zou, and Ruijie Xu for their discussion on the method prototype.
We also appreciate Jiadi Su, Kang Xu, Minyue Dai, Rui Li, Siyu Lu, Shijie Xia, Tang Tang, and Yanan Wang for their contribution in the experiment.
This project is supported by Qingyuan Research Project and Shanghai Artificial Intelligence Laboratory.

\bibliography{custom}

\clearpage
\appendix
\section{More Extensive Evaluation}
\label{sec:more_evaluation}
\subsection{Human Evaluation on Story Premises}
\label{sec:appendix_human_eval_premise}
We set up a human evaluation study on story premises generated by MoPS and baselines.
Specially, we enlisted four evaluators (two men and two women) who were not previously involved with our project.
This group consisted of one PhD student with significant AI expertise, two early-stage PhD students, and one person outside academia.

We chose the three highest-scoring baselines for comparison alongside our method: Complex (CPX), the top LLM-based baseline; Storium (STM), the leading source from existing public datasets; and WritingPrompts (WP), the most frequently used in past studies.
We took 20 story premises from each method for evaluation.

Consistent with the quality metrics used in \texttt{gpt-4-turbo} evaluation (Tab.~\ref{tab:premise_scores}), we asked evaluators to rate each premise on Fascination, Completeness, and Originality on a 1 to 5 scale, requiring them to review 20 * 4 * 3 = 240 items in total. The results were gathered via a survey, highlighting any statistically significant differences ($p$ < 0.05):
\begin{table}[htbp]
\centering
\resizebox{\columnwidth}{!}{
\begin{tabular}{@{}lccccc@{}}
\toprule
 & Fascination$\uparrow$ & Completeness$\uparrow$ & Originality$\uparrow$ & Average$\uparrow$ \\
\midrule
MoPS & \textbf{3.4125} & \textbf{4.0375} & \textbf{3.2375} & \textbf{3.5625} \\
CPX & \textbf{3.125} & \textbf{3.8875} & 2.75 & 3.2542 \\
STM & 2.8875 & 2.9875 & \textbf{3.3375} & 3.0708 \\
WP & \textbf{3.0875} & 3.05 & \textbf{3.175} & 3.1042 \\
\bottomrule
\end{tabular}}
\caption{Human evaluation results on story premises. The optimal values (with $p$ < 0.05) are \textbf{bolded}.}
\label{tab:scores}
\end{table}

\subsection{Claude-3 Evaluation on Story Premsies and Premise-based Stories}

For assessing long stories like scripts and novels created from premises, we considered the need for human evaluators to review more than 100,000 words across 20 samples, making it challenging for them to maintain high-quality assessments. Indeed, most evaluators also decline to review such extensive materials.
As a result, we opted to use \texttt{claude-3-opus}, the most advanced LLM available, to assess the premises and the resulting scripts and novels.

We chose 100 premises from the four methods discussed above, 30 novels generated using RecurrentGPT~\cite{DBLP:journals/corr/abs-2305-13304}, and 30 scripts generated with Dramatron~\cite{DBLP:conf/chi/MirowskiMPE23} for this evaluation. The following are the results:
\begin{table}[htbp]
\centering
\resizebox{\columnwidth}{!}{
\begin{tabular}{@{}lccccc@{}}
\toprule
 & Fascination$\uparrow$ & Completeness$\uparrow$ & Originality$\uparrow$ & Average$\uparrow$ \\
\midrule
MoPS & \textbf{73.65} & \textbf{72.35} & \textbf{94.75} & \textbf{80.25} \\
CPX & 71.22 & 66.40 & 84.65 & 74.09 \\
STM & \textbf{73.66} & 67.40 & 89.65 & 76.90 \\
WP & 70.74 & 51.90 & \textbf{93.70} & 72.11 \\
\bottomrule
\end{tabular}}
\caption{\texttt{Claude-3-opus} evaluation results on story premises. The optimal values (with $p$ < 0.05) are \textbf{bolded}.}
\label{tab:claude3_scores}
\end{table}

\begin{table}[htbp]
    \centering
    \small
    \resizebox{\columnwidth}{!}{
    \begin{tabular}{@{}lcccc@{}}
    \toprule
    & Fascination$\uparrow$ & Completeness$\uparrow$ & Originality$\uparrow$ & Average$\uparrow$\\ \midrule
    MoPS-RecurrentGPT & \textbf{73.00} & \textbf{64.67} & \textbf{85.00} & \textbf{74.22} \\
    CPX-RecurrentGPT & 71.47 & \textbf{64.17} & \textbf{85.00} & \textbf{73.54} \\
    STM-RecurrentGPT & 69.20 & 57.83 & \textbf{85.00} & 70.79 \\
    WP-RecurrentGPT & 71.13 & \textbf{62.67} & \textbf{85.33} & \textbf{72.83} \\
    \midrule
    MoPS-Dramatron & \textbf{78.97} & \textbf{82.50} & \textbf{86.17} & \textbf{82.54} \\
    CPX-Dramatron & \textbf{78.33} & \textbf{81.83} & \textbf{86.17} & \textbf{82.11} \\
    STM-Dramatron & 73.50 & 76.17 & \textbf{82.83} & 77.50 \\
    WP-Dramatron & 72.76 & 73.33 & \textbf{82.17} & 76.09 \\
    \bottomrule
    \end{tabular}}
    \caption{\texttt{Claude-3-opus} evaluation results on premise-based stories. The optimal values (with $p$ < 0.05) are \textbf{bolded}.}
    \label{tab:claude3_story_score}
\end{table}

The evaluations by humans and Claude-3 show that the quality of story premises made by MoPS matches the GPT-4 assessment results in \S\ref{sec:exp_results}. Claude-3 also found that the strengths of MoPS premises carry over to the extended stories, agreeing with GPT-4’s views and supporting the assessments we discussed.

\section{Further Study on Reliability of Automatic Diversity Metrics}
\label{sec:further_study_diversity_metrics}
\subsection{Robustness of Different Settings}

For diversity breadth and density, using different t-SNE settings and SentenceBert models can produce slightly different polygons, leading to slightly varying results.
This might make one wonder about the consistency of our measures.
To tackle this issue, we conducted the following experiment:

\noindent\textbf{Different t-SNE random states}. We used 5 random seeds for t-SNE and then calculated the breadth and density of these using Eq.~\ref{eq:breadth} and~\ref{eq:density}. Below, we share the average results from these five different trials in Tab.~\ref{tab:tSNE_different_seeds} (left half).

\noindent\textbf{Different SentenceBert models}. We selected 3 different SentenceBert models: \textit{all-MiniLM-L6-v2} (used in the paper), \textit{all-mpnet-base-v2}, and \textit{all-MiniLM-L12-v2}. Tab.~\ref{tab:tSNE_different_seeds} (right half) reports the average score for these three models:
\begin{table}[htbp]
\centering
\small
\resizebox{\columnwidth}{!}{
\begin{tabular}{@{}lrrrrr@{}}
\toprule
\multirow{2}{*}{} &  \multicolumn{2}{c}{t-SNE random states} & \multicolumn{2}{c}{SentenceBert models}\\ \cmidrule(l){2-3} \cmidrule(l){4-5} 
 & Breadth$\uparrow$ & Density$\downarrow$ &  Breadth$\uparrow$ & Density$\downarrow$ & \\
\midrule
MoPS & \textbf{3389.3868} & \textbf{8.0092} & \textbf{3430.269} & \textbf{8.524} \\
CPX & 2664.791 & 14.957 & 2730.924 & 12.640 \\
VIL & 3089.7938 & 11.1426 & 3100.338 & 11.127 \\
DOC & 1970.926 & 13.7998 & 1948.131 & 14.698 \\
WP & 3069.2838 & 8.7834 & 2949.854 & 8.881 \\
STM & 1964.940 & 14.6374 & 1795.277 & 15.957 \\
\bottomrule
\end{tabular}}
\caption{Average diversity score of five different \texttt{random\_state} of t-SNE (left half) and three different SentenceBert models (right half).}
\label{tab:tSNE_different_seeds}
\end{table}

The results above indicate that variations in t-SNE hyperparameters and changes in SentenceBert models do not affect the superiority of MoPS in terms of semantic diversity (breadth and density).

\subsection{Alignment with Human Intuition}

To verify if the semantic breadth and depth experiments proposed in our paper align with human intuition, we organized a human evaluation experiment.
Specifically, we selected 100 story premises from MoPS and five other baselines, putting each group's 100 premises on a single page of a questionnaire, creating a 6-page document.
We then asked human evaluators to read all 600 story premises, 100 from each group, and rate each group's semantic diversity based on their intuition and instinct, using a scale from 1 to 5. We specifically instructed evaluators to differentiate their scores and avoid giving a score of 3 as much as possible.

For the human evaluators, we brought back the 4 evaluators mentioned in \S\ref{sec:appendix_human_eval_premise} and added 4 more (two men and two women). This new group included a senior master's student with several publications, a senior undergraduate, and two non-researchers. Tab.~\ref{tab:align_with_human_intuition} reports the evaluation results:
\begin{table}[h!]
\centering
\resizebox{\columnwidth}{!}{
\begin{tabular}{@{}lrrrrrrrrr@{}}
\toprule
  & Average & E1* & E2* & E3* & E4 & E5 & E6 & E7* & E8 \\
\midrule
MoPS & 3.875 & 4 & 5 & 3 & 4 & 4 & 3 & 4 & 4 \\
CPX & 2.25 & 4 & 2 & 2 & 2 & 3 & 1 & 1 & 3 \\
VIL & 2.625 & 3 & 2 & 4 & 3 & 3 & 2 & 2 & 2 \\
DOC & 3.5 & 3 & 2 & 4 & 5 & 5 & 4 & 3 & 4 \\
WP & 3.75 & 2 & 5 & 5 & 2 & 5 & 5 & 4 & 2 \\
STM & 3.125 & 3 & 3 & 3 & 3 & 4 & 4 & 2 & 2 \\
\bottomrule
\end{tabular}}
\caption{Human evaluation results (E $\rightarrow$ Evaluator, * indicates evaluators from \S\ref{sec:appendix_human_eval_premise}). Eight evaluators provide intuitive judgments on the diversity of story premises.}
\label{tab:align_with_human_intuition}
\end{table}

The results showed that MoPS and WP had the highest scores, which aligns with the diversity scores presented in Tab.~\ref{tab:premise_scores}. Interestingly, while MoPS received fewer top scores than WP, its scores were more consistently high across all eight evaluators. We also recognize that the high cost of human evaluation makes it hard to obtain results with significant differences, which is a limitation of human assessments. Despite this, we think the human evaluation experiment backs up our diversity metrics as being in line with human intuition.

\section{Baseline Details}
\label{appendix: baselines}

We provide prompts and examples for each baseline in Tab.~\ref{tab:baselines_case_study}.

\textbf{Vanilla (VIL)}: instruct \texttt{gpt-3.5-turbo} to generate premises with temperature=0.6.
We use the prompt shown in Tab.~\ref{tab:baselines_case_study} to generate 10 story premises at a time.
Whenever a new permise is generated, we deduplicated based on the cosine similarity of sentence embeddings~\cite{DBLP:conf/emnlp/ReimersG19}, excluding items with a threshold $\epsilon\ge0.85$.

\textbf{Complex (CPX)}: similar to VIL, but with 3 premises synthesized by MoPS as few-shot examples integrated into the prompt. 
The purpose of this baseline is to explore whether LLM can produce comparable story premises when provided with high-quality story premises as few-shot examples.

\textbf{DOC}: originally stemming from~\citet{DBLP:journals/corr/abs-2310-08796}, their research explored the feasibility of end-to-end story plot generation.
They instructed \texttt{llama2-13b-chat}~\cite{DBLP:journals/corr/abs-2307-09288} to write 7,000 story premises via prompt: \textit{``Write a premise for a short story.''} and paired each with two story plots generated by oasst-30b~\cite{DBLP:journals/corr/abs-2304-07327}.
After being curated by original authors, it was publicly released in doc-story-gen-v2\footnote{\url{https://github.com/facebookresearch/doc-storygen-v2}} repository.
The purpose of this baseline is to explore the story premise generation capability of open-source LLMs.
We randomly extracted 1,000 entries for evaluation.

\textbf{WritingPrompts (WP)}: collected by ~\citet{DBLP:conf/acl/LewisDF18} from Reddit's writingPrompts forum, it includes approximately 300k story premises and corresponding short stories written by human.
A significant amount of research work~\cite{DBLP:conf/naacl/TanYAXH21,DBLP:conf/icml/PapalampidiCK22,DBLP:conf/naacl/HanCTP22,DBLP:conf/coling/SunSMLF22,li2023enhancing,DBLP:journals/corr/abs-2310-08185} has utilized these story premises to validate their methods.

\textbf{Storium (STM)}: released by~\citet{DBLP:conf/emnlp/AkouryWWHPI20}. They collected 5,743 publicly available stories from the turn-based role-playing game platform - STORIUM\footnote{\url{https://storium.com/}}.
It requires a small group of human users to collaborate on a card game.
All settings of the storytelling game are served as a highly structured story.
We use the description of each storytelling game as the story premise.

\section{Detailed Experiment Results}
\label{appendix:detailed_exp_results}
\subsection{Evaluation Results on Premise Diversity}
\label{appendix:diversity_fig}
\begin{figure*}[ht] \centering
    \includegraphics[width=\textwidth]{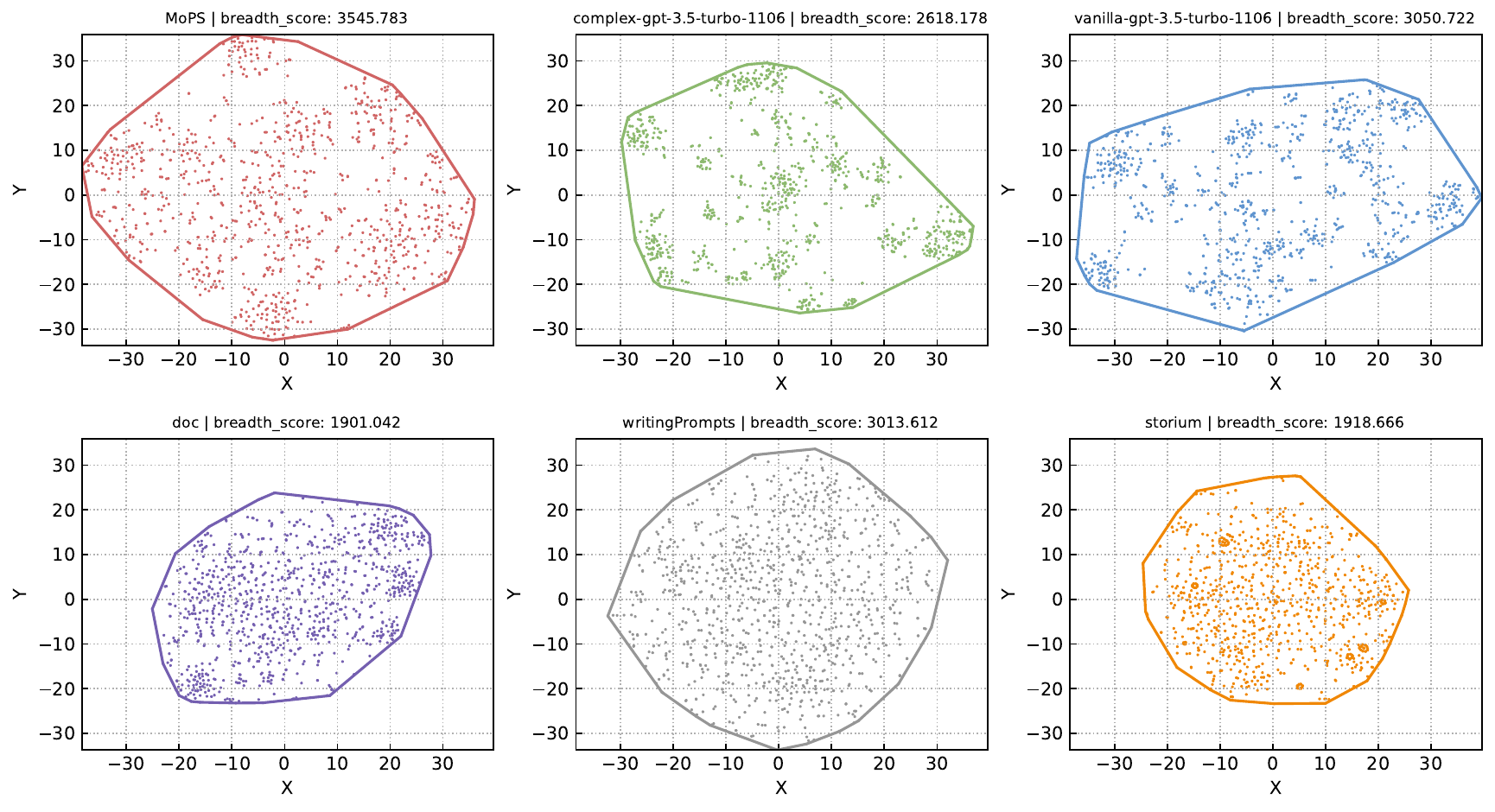}
    \vspace{-0.5cm}
    \caption{Breadth score of all methods.
    The premises synthesized by MoPS surpassed comparative methods in semantic breadth.
    \textcolor{red}{Note}: Chrome or Edge browser may not display this figure properly. Please use a specialized PDF viewer.
    } \label{fig:breadth_score}
\end{figure*}

Fig.~\ref{fig:breadth_score} and~\ref{fig:density_score} show 
the diversity score for all methods in terms of semantic breadth and density.

\subsection{Evaluation Results on Premise Quality}
\label{appendix:quality_fig}
\begin{figure*}[ht] \centering
    \includegraphics[width=\textwidth]{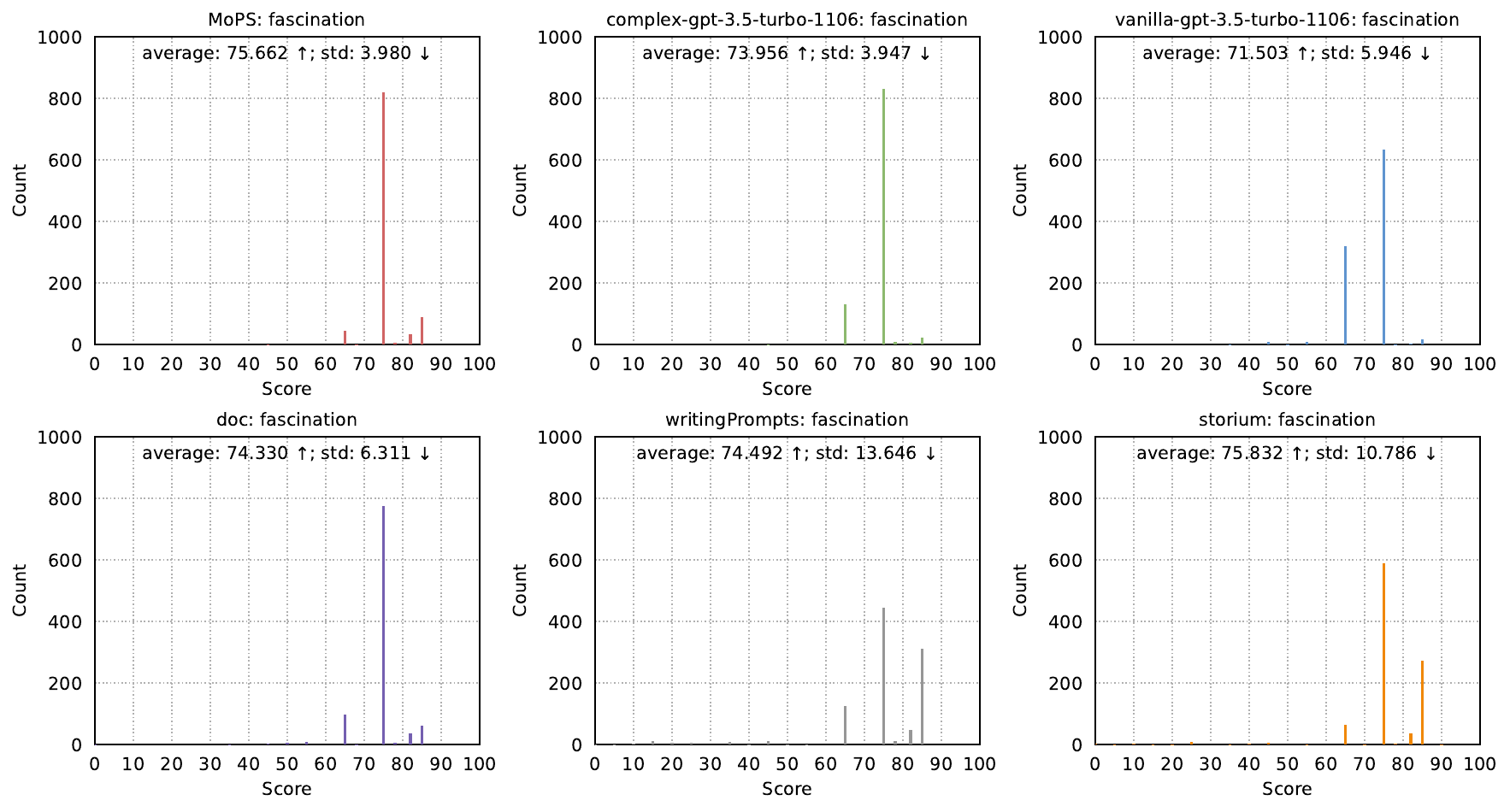}
    \vspace{-0.5cm}
    \caption{Fascination score of all methods. 
    MoPS's average score is superior to all baselines except for storium, and it has the lowest score standard deviation.
    This indicates that the premises synthesized by MoPS are appealing and of stable quality. 
    Additionally, the story premises collected by Storium require a group of people to participate in a collaborate game, whereas MoPS is fully applicable in situations without human participation.} \label{fig:fascination_score}
\end{figure*}

\begin{figure*}[ht] \centering
    \includegraphics[width=\textwidth]{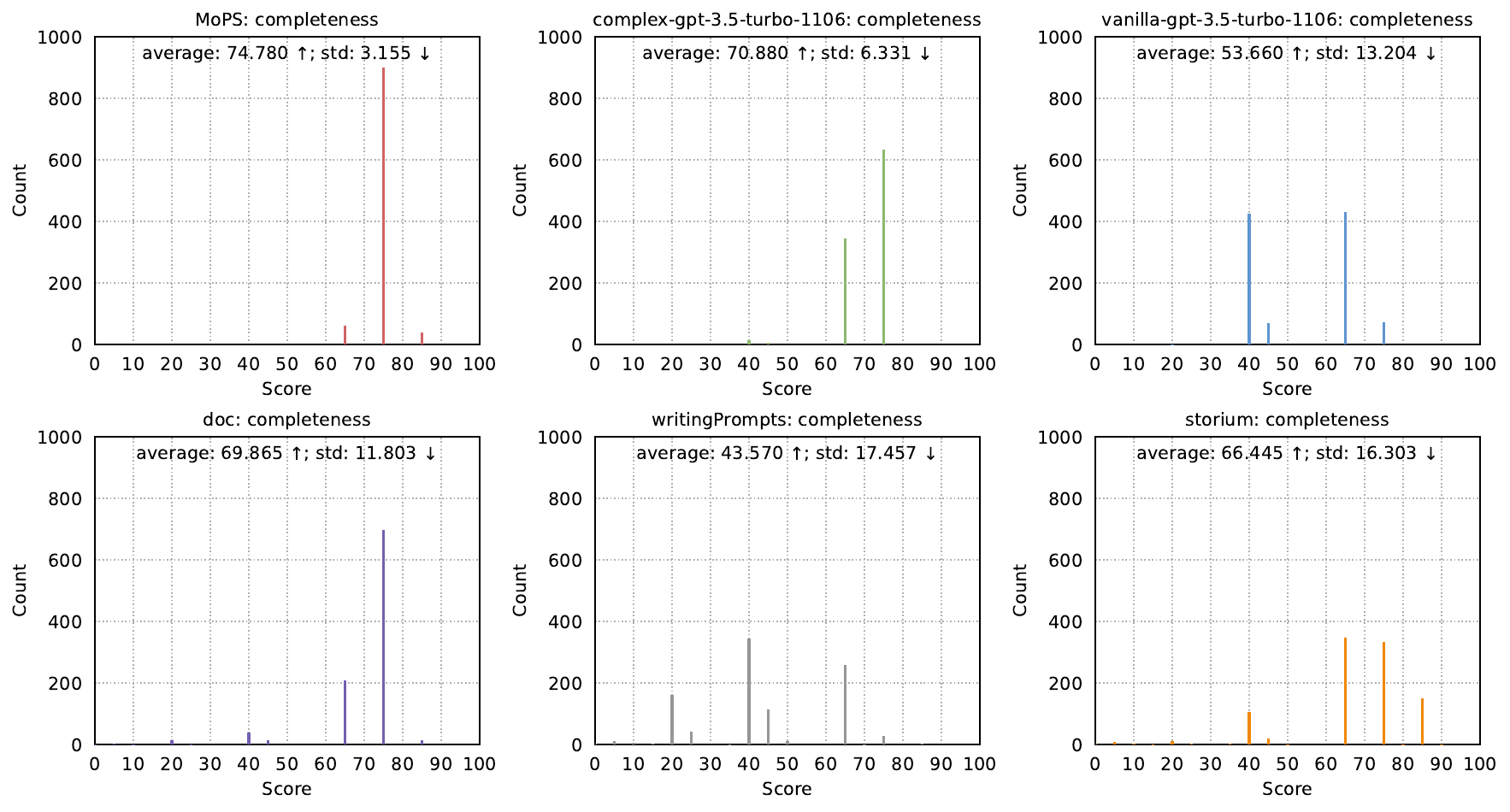}
    \vspace{-0.5cm}
    \caption{Completeness score of all methods.
    MoPS's average score and standard deviation are superior to all baselines.
    Particularly, the completeness of two human-written datasets is much lower than MoPS and has much higher standard deviations, which indicates instability of premise quality within the dataset.
    } \label{fig:completeness_score}
\end{figure*}

\begin{figure*}[ht] \centering
    \includegraphics[width=\textwidth]{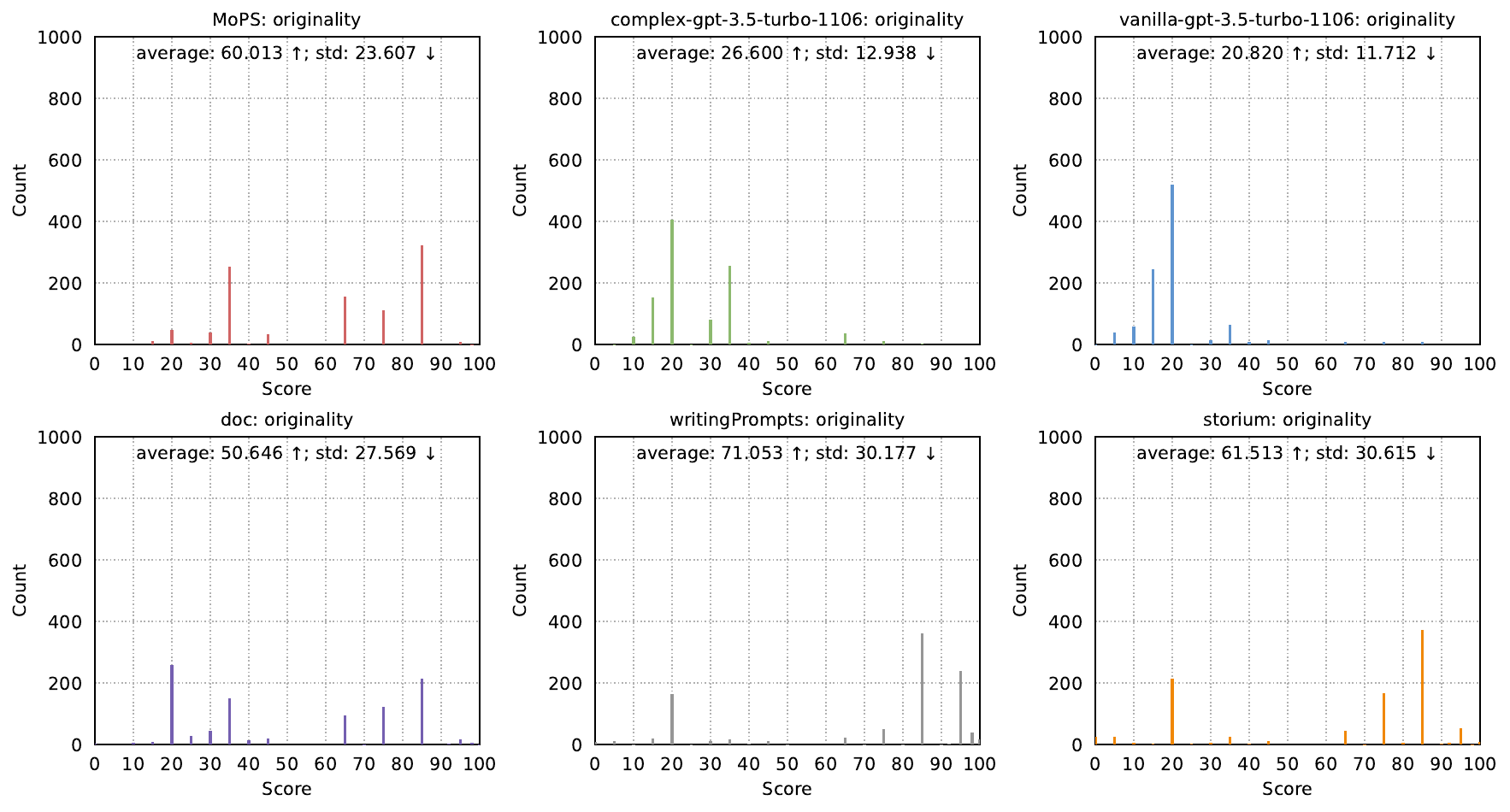}
    \vspace{-0.5cm}
    \caption{Originality score for all methods. Apart from WP and STM, MoPS's average score surpasses all other baselines. Since all components in MoPS still originate from LLM, compared to premises entirely written by humans from WP and STM, MoPS is slightly inferior in originality.
    However, it is surprising that MoPS shows a significant improvement in originality compared to VIL and CPX.
    This is attributed to the combinatorial 
creativity of modules, leading to unique and innovative outcomes.} \label{fig:originality_score}
\end{figure*}

Fig.~\ref{fig:fascination_score},~\ref{fig:completeness_score},~\ref{fig:originality_score} show the distribution, average, and standard deviation of fascination, completeness, and originality scores for all methods, respectively.

\section{Curated Dataset}
\label{appendix:curated_dataset}
\begin{figure*}[ht] \centering
    \includegraphics[width=\textwidth]{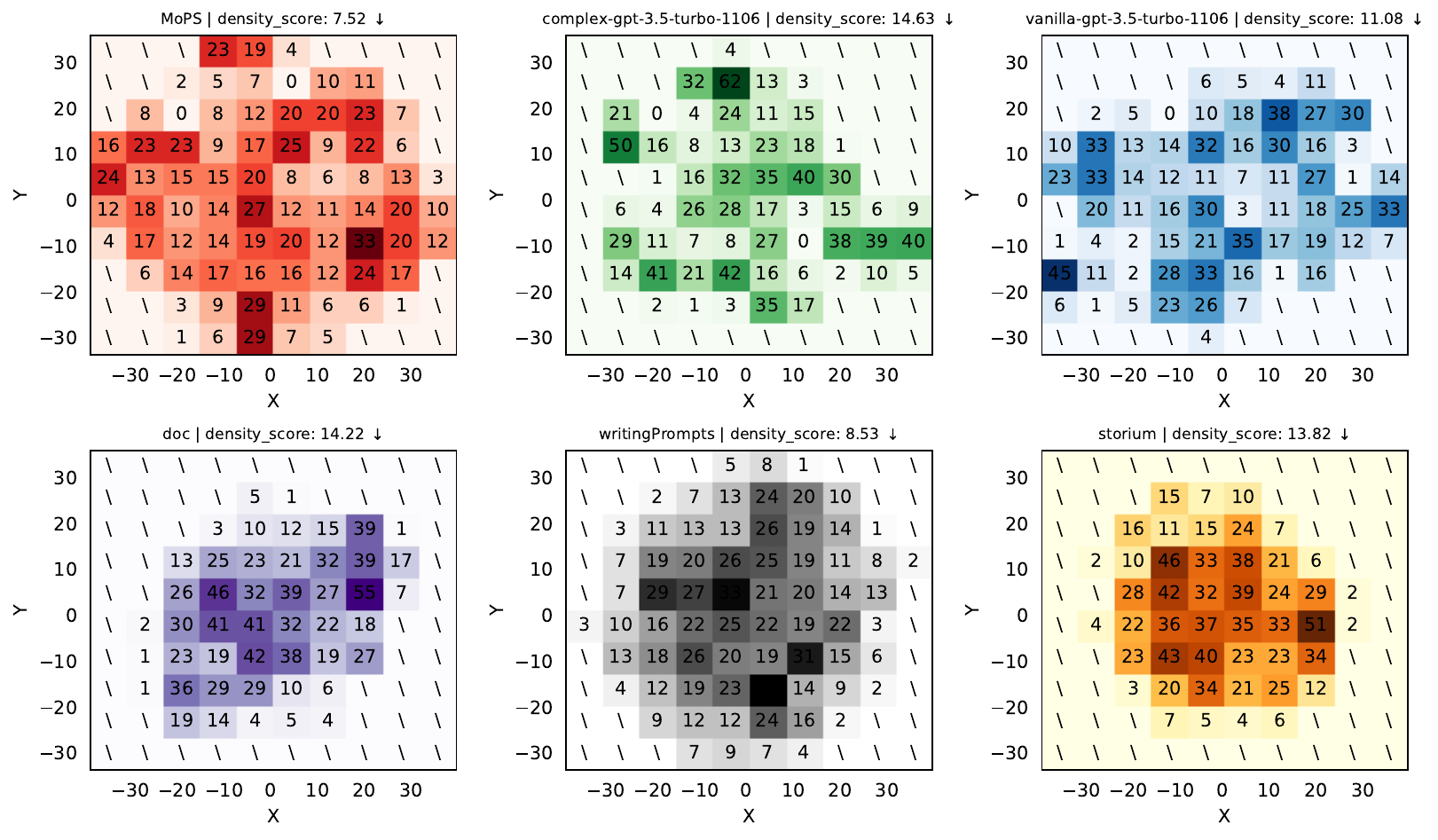}
    \vspace{-0.5cm}
    \caption{Density score of all methods.   %
    The premises synthesized by MoPS surpassed comparative methods in semantic density.} \label{fig:density_score}
\end{figure*}

The purpose is to collect \textbf{high-quality} and \textbf{diverse} story premises from the moderate version of the dataset to form a curated dataset.
We draw inspiration from a classic method in Quality-Diversity field~\cite{DBLP:journals/firai/PughSS16}: Map-Elites~\cite{DBLP:journals/corr/MouretC15} to meticulously craft the dataset.
Specifically, within the semantic 2D histogram of MoPS (see Fig.~\ref{fig:density_score}), there are a total of \textbf{74} bins that are both valid and have a count greater than 0.
We select the story premise with the highest total quality score (fascination score + completeness score + originality score) from each bin.
For the the rest of entries, we rank the remaining 926 story premises in moderate dataset by total quality score and choose the top \textbf{26} entries.
Finally, we extract the novels and scripts paired with these 100 story premises to form the curated dataset.

\section{Prompts used in Story Evaluation}
\label{subsec:prompts_in_story_evaluation}
Tab.~\ref{tab:prompt_for_story_fascination}, Tab.~\ref{tab:prompt_for_story_completeness}, and Tab.~\ref{tab:prompt_for_story_originality} are prompts for fascination, completeness and originality score used in story evaluation.

\section{Prompts used in MoPS}  
\label{sec:appendix_prompts}
Tab.~\ref{tab:prompt_for_inducing_backgrounds} is the  prompt for inducing backgrounds. The {component} is one of three: time, place, or time and place.
Tab.~\ref{tab:prompt_for_inducing_persona} is the prompt for inducing personas, including three categories: growth, conflict, cooperation.
Tab.~\ref{tab:prompt_for_inducing_events},~\ref{tab:prompt_for_inducing_endings},~\ref{tab:prompt_for_inducing_twists} are respectively prompt for inducing events, endings, and twists.

Tab.~\ref{tab:prompt_for_synthesis} and~\ref{tab:prompt_for_verification} are prompts used for synthesizing and verifying premise, respectively.

\section{Prompts used in Premise Evaluation}
\label{appendix: quality_prompts}
Tab.~\ref{tab:prompt_for_fascination}, Tab.~\ref{tab:prompt_for_completeness}, and Tab.~\ref{tab:prompt_for_originality} are prompts for fascination, completeness and originality score used in premise evaluation.

\section{Example of Premise Design}
Tab.~\ref{tab:examples_of_design_of_premise} shows the manually pre-defined theme candidates and an example of premise design within the collected nested dictionary.
We will release the \textbf{code} for MoPS as well as \textbf{all premise designs} collected from \texttt{gpt-3.5-turbo} used in this paper (essentially a nested dictionary).

\section{Example of Premise-Based Story}
\label{appendix:examples_of_novels_and_scripts}

Tab.~\ref{tab:script_and_novel} shows a example of script and novel generated from a MoPS premise by Dramatron\footnote{\url{https://github.com/google-deepmind/dramatron}} and RecurrentGPT\footnote{\url{https://github.com/aiwaves-cn/RecurrentGPT}} respectively.
All stories are carried out with \texttt{gpt-3.5-turbo} as the language backend.
The moderate version dataset contains a total of 1,000 such novels and scripts, which will be publicly released to contribute to the field of automatic story generation.


\clearpage

\begin{table*}[htb]
    \centering
    \small
    \begin{tabular}{>{\raggedright\arraybackslash\tt}p{0.98\textwidth}<{}}
        \toprule
        Tell me 10 backgrounds in \textcolor{gray}{\{theme\}} themed novels and scripts. \\ 
        \\
        Each background should only include \textcolor{gray}{\{component\}} behind literary works and no any other extra narratives. \\
        \\
        Each line starts with a serial number and a dot. \\
        \bottomrule
    \end{tabular}
    \caption{Prompt for inducing backgrounds. The \{component\} is one of three: time, place, or time and place.}
    \label{tab:prompt_for_inducing_backgrounds}
\end{table*}
\begin{table*}[htb]
    \centering
    \small
    \begin{tabular}{>{\raggedright\arraybackslash\tt}p{0.98\textwidth}<{}}
        \toprule
        \textcolor{green}{Growth (only include a protagonist)}\\
        \midrule
            The following is the theme and background of a novel or script:\\
            \themeprompt
            \bgprompt
            \\
            Based on the theme and background mentioned above, tell me 3 possible protagonists.\\
            The protagonist is the main character portrayed in the narratives about their growth.\\
            Each protagonist should only include a brief characterization, without specific names.\\
            Each output line starts with a serial number and a dot. \\
        \midrule
        \textcolor{green}{Conflict (include a protagonist and an antagonist)}\\
        \midrule
            The following is the theme and background of a novel or script:\\
            \themeprompt
            \bgprompt
            \\
        Based on the theme and background mentioned above, tell me 3 possible (protagonist, antagonist) .
        \\
        The protagonist is the main character portrayed in the narratives about their growth.
        \\
        The main role of the antagonist is to create a conflict event with the protagonist to prevent it from achieving its goal.
        \\
        Each pair should be presented in the format: protagonist: <a brief characterization>; antagonist: <a brief characterization>.
        \\
        Each output line starts with a serial number and a dot, contains a (protagonist, antagonist) pair.
        \\
        Please remember to use protagonist and antagonist without specific names appearing. \\
        \midrule
        \textcolor{green}{Cooperation (include a protagonist and a deuteragonist)}\\
        \midrule
            The following is the theme and background of a novel or script:\\
            \themeprompt
            \bgprompt
            \\
        Based on the theme and background mentioned above, tell me 3 possible (protagonist, deuteragonist) . \\
        The protagonist is the main character portrayed in the narratives about their growth. \\
        The main role of the deuteragonist is to collaborate with the protagonist to achieve its goal. \\
        Each pair should be presented in the format: protagonist: <a brief characterization>; deuteragonist: <a brief characterization>. \\
        Each output line starts with a serial number and a dot, contains a (protagonist, deuteragonist) pair. \\
        Please remember to use protagonist and deuteragonist without specific names appearing. \\
        \bottomrule
    \end{tabular}
    \caption{Prompt for inducing personas, including three categories: growth, conflict, cooperation.}
    \label{tab:prompt_for_inducing_persona}
\end{table*}
\begin{table*}[htb]
    \centering
    \small
    \begin{tabular}{>{\raggedright\arraybackslash\tt}p{0.98\textwidth}<{}}
        \toprule
            The following is the theme, background and persona of a novel or script:\\
            \themeprompt
            \bgprompt
            \personaprompt
            \\
            Based on the theme, background and persona mentioned above, conceive two independent events that could run through the entire narrative context. \\
            \\
            Please use a concise and coherent sentence to describe the entire event.\\
        \bottomrule
    \end{tabular}
    \caption{Prompt for inducing events.}
    \label{tab:prompt_for_inducing_events}
\end{table*}
\begin{table*}[ht]
    \centering
    \small
    \begin{tabular}{>{\raggedright\arraybackslash\tt}p{0.98\textwidth}<{}}
        \toprule
            The following is the theme, background, persona and main event of a novel or script:
            \\
            \themeprompt
            \bgprompt
            \personaprompt
            \eventprompt
            \\
            Based on the theme, background, persona and event mentioned above, conceive an concretized ending. \\
            \\
            Please use a concise and coherent sentence to describe the ending.\\
        \bottomrule
    \end{tabular}
    \caption{Prompt for inducing endings.}
    \label{tab:prompt_for_inducing_endings}
\end{table*}
\begin{table*}[ht]
    \centering
    \small
    \begin{tabular}{>{\raggedright\arraybackslash\tt}p{0.98\textwidth}<{}}
        \toprule
            The following is the theme, background, persona, main event and ending of a novel or script:
            \\
            \themeprompt
            \bgprompt
            \personaprompt
            \eventprompt
            \finalendingprompt
            \\
            Based on the theme, background, persona, event and ending mentioned above, conceive a twist as an unique hook to connect the main event and ending. \\
            \\
            Please use a concise and coherent sentence to describe the twist.\\
        \bottomrule
    \end{tabular}
    \caption{Prompt for inducing twists.}
\label{tab:prompt_for_inducing_twists}
\end{table*}
\begin{table*}[ht]
    \centering
    \small
    \begin{tabular}{>{\raggedright\arraybackslash\tt}p{0.98\textwidth}<{}}
        \toprule
            The following is the theme, background, persona, main event, final ending and twist of a novel or script:
            \\
            \themeprompt
            \bgprompt
            \personaprompt
            \eventprompt
            \finalendingprompt
            \twistprompt
            \\
            Please combine the aforementioned elements of a novel or script into one compact, concise, and coherent sentence as a story premise. \\
        \bottomrule
    \end{tabular}
    \caption{Prompt for premise synthesis.}
\label{tab:prompt_for_synthesis}
\end{table*}
\begin{table*}[ht]
    \centering
    \small
    \begin{tabular}{>{\raggedright\arraybackslash\tt}p{0.98\textwidth}<{}}
        \toprule
            Here is a story premise: \\
            \\
            \textcolor{purple}{\{premise\}} \\
            \\
            Please help to verify: \\
            \\
            1. Does it contain obvious inconsistencies. For example, the background, plot, and characters do not match. \\
            \\
            2. Does it contain obvious factual errors. For example, there were obvious historical errors and time span errors. \\ 
            \\
            If there are any errors mentioned above, please return Yes wrapped by [[]], otherwise return No wrapped by [[]] without any other extra output. \\
        \bottomrule
    \end{tabular}
    \caption{Prompt for premise verification.}
\label{tab:prompt_for_verification}
\end{table*}

\begin{table*}[ht]
    \centering
    \small
    \begin{tabular}{>{\raggedright\arraybackslash\tt}p{0.98\textwidth}<{}}
        \toprule
            Here is a story premise: \\
            \\
            \textcolor{purple}{\{premise\}} \\
            \\
            Now let you give a score from 0 to 100 to assess to its fascination. \\
            \\
            Score 0 indicates that this premise is completely confused, while score 100 indicates that you really want to see the story created based on this premise. \\
            \\
            Requirement: just provide a deterministic score and provide a concise and brief explanation, with a blank line between the two.\\
            \\
            Score:\\
        \bottomrule
    \end{tabular}
    \caption{Prompt for fascination score used in premise evaluation.}
\label{tab:prompt_for_fascination}
\end{table*}
\begin{table*}[ht]
    \centering
    \small
    \begin{tabular}{>{\raggedright\arraybackslash\tt}p{0.98\textwidth}<{}}
        \toprule
            Here is a story premise: \\
            \\
            \textcolor{purple}{\{premise\}} \\
            \\
            Now let you give a score from 0 to 100 which represents its completeness level. \\
            \\
            Score 0 indicates that it lacks all elements , while score 100 indicates that it has all elements. \\
            \\
            Requirement: just provide a deterministic score and provide a concise and brief explanation, with a blank line between the two. \\
            \\
            Score:\\
        \bottomrule
    \end{tabular}
    \caption{Prompt for completeness score used in premise evaluation.}
\label{tab:prompt_for_completeness}
\end{table*}
\begin{table*}[ht]
    \centering
    \small
    \begin{tabular}{>{\raggedright\arraybackslash\tt}p{0.98\textwidth}<{}}
        \toprule
            Here is a story premise: \\
            \\
            \textcolor{purple}{\{premise\}} \\
            \\
            Now let give you a score from 0 to 100 which represents your level of familiarity with it. \\
            \\
            Score 0 indicates that you have seen the exact same premise, while score 100 indicates that you have never seen the same premise at all. \\
            \\
            Your score should be based on the assumption that the candidate is at least a complete story premise. Otherwise, you should give a score 0. \\
            \\
            Requirement: just provide a deterministic score and provide a concise and brief explanation, with a blank line between the two.\\
           \\
            Score:\\
        \bottomrule
    \end{tabular}
    \caption{Prompt for originality score used in premise evaluation.}
\label{tab:prompt_for_originality}
\end{table*}

\begin{table*}[ht]
    \centering
    \small
    \begin{tabular}{>{\raggedright\arraybackslash\tt}p{0.98\textwidth}<{}}
        \toprule
            Here is a \{story\_type\}: \\
            \\
            \textcolor{purple}{\{story\}} \\
            \\
            Now let you give a score from 0 to 100 to assess to its fascination. \\
            \\
            Score 0 indicates that the \{story\_type\} is completely confused, while score 100 signifies that the \{story\_type\} is bound to become a worldwide sensation. \\
            \\
            Requirement: just provide a deterministic score and provide a concise and brief explanation, with a blank line between the two.\\
            \\
            Score:\\
        \bottomrule
    \end{tabular}
    \caption{Prompt for fascination score used in story evaluation. The \{story\_type\} is one of two: novel and script.}
\label{tab:prompt_for_story_fascination}
\end{table*}
\begin{table*}[ht]
    \centering
    \small
    \begin{tabular}{>{\raggedright\arraybackslash\tt}p{0.98\textwidth}<{}}
        \toprule
            Here is a \{story\_type\}: \\
            \\
            \textcolor{purple}{\{story\}} \\
            \\
            Now let you give a score from 0 to 100 which represents its completeness level. \\
            \\
            Score 0 indicates that it lacks all elements a \{story\_type\} should have, while score 100 indicates that it has all elements a \{story\_type\} should have.\\
            \\
            Requirement: just provide a deterministic score and provide a concise and brief explanation, with a blank line between the two. \\
            \\
            Score:\\
        \bottomrule
    \end{tabular}
    \caption{Prompt for completeness score used in story evaluation. The \{story\_type\} is one of two: novel and script.}
\label{tab:prompt_for_story_completeness}
\end{table*}
\begin{table*}[ht]
    \centering
    \small
    \begin{tabular}{>{\raggedright\arraybackslash\tt}p{0.98\textwidth}<{}}
        \toprule
            Here is a \{story\_type\}: \\
            \\
            \textcolor{purple}{\{story\}} \\
            \\
            Now let give you a score from 0 to 100 which represents your level of familiarity with it. \\
            \\
            Score 0 indicates that you have seen \{story\_type\}s that are very similar to as the one provided, while score 100 means that you have never seen a \{story\_type\} that is very similar to the one provided. \\
            \\
            Requirement: just provide a deterministic score and provide a concise and brief explanation, with a blank line between the two.\\
           \\
            Score:\\
        \bottomrule
    \end{tabular}
    \caption{Prompt for originality score used in story evaluation. The \{story\_type\} is one of two: novel and script.}
\label{tab:prompt_for_story_originality}
\end{table*}

\begin{table*}[htb]
    \centering
    \small
    \begin{tabular}{>{\raggedright\arraybackslash\tt}p{0.98\textwidth}<{}}
        \toprule
        \textbf{Prompt for Vanilla Baseline} \\
        \midrule
        Write 10 premises for novels or scripts in one sentence. \\
        \midrule
        \textbf{Prompt for Complex Baseline} \\
        \midrule
        Write 10 premises for novels or scripts in one sentence like below. \\
        \\
        \{3 premises synthesized by MoPS\} \\
        \midrule
        \textbf{A premise entry in DOC dataset} \\
        \midrule
        In a dystopian future where emotions are outlawed, a young man named Leo discovers he is still capable of feeling love. Tormented by his forbidden affection, Leo must decide whether to face the consequences of living in a society that punishes any display of affection or run from his oppressive government. \\
        \midrule
        \textbf{A premise entry in writingPrompts dataset} \\
        \midrule
        $[ \text{WP} ]$ You 've finally managed to discover the secret to immortality. Suddenly, Death appears before you, hands you a business card, and says, When you realize living forever sucks, call this number , I 've got a job offer for you. \\
        \midrule
        \textbf{A premise entry in storium dataset} \\
        \midrule
        Smooth seas, cool breeze, clear moonlight. The night was going so peacefully up until that whole sneak attack business. Funny, you don't remember making any new enemies lately. But somebody obviously has it in for you  nobody hunts you down like that for no reason. Who is it? And what in the world do they have against you, anyway? \\
        \bottomrule
    \end{tabular}
    \caption{Prompts for vanilla and complex baseline, and an example premise entry in DOC, wringtPrompts, and storium dataset.}
    \label{tab:baselines_case_study}
\end{table*}

\begin{table*}[htb]
    \centering
    \small
    \begin{tabular}{>{\raggedright\arraybackslash\tt}p{0.98\textwidth}<{}}
        \toprule
        \textbf{Pre-defined Theme Candidates} \\
        \midrule
        \textcolor{gray}{Historical, Game, Time-travel, Immortal Heroes, Contemporary, Suspense, Sports, Fantastic, Science Fiction, Martial Arts, Military, Urban, Romance, Fantasy} \\
        \midrule
        \textbf{A Design of Premise in Collected Nested Dictionary} \\ 
        \midrule
\begin{lstlisting}[language=json,firstnumber=1, escapeinside={(*@}{@*)}]
(*@ \textcolor{gray}{Theme} $\triangleright$ @*)"Fantasy": {
(*@ \textcolor{red}{Background} $\triangleright$ @*)"A medieval kingdom in the land of Eldoria. ": {
              "conflict": {
(*@ \textcolor{green}{Persona} $\triangleright$ @*)       "Protagonist: A young, skilled knight seeking to prove their worth              and protect the kingdom; Antagonist: A power-hungry sorcerer who              seeks to overthrow the kingdom and rule with dark magic.": {
(*@ \textcolor{blue}{Plot:Event} $\triangleright$ @*)        "The protagonist embarks on a quest to retrieve a powerful                  artifact that can help defeat the antagonist and save the                  kingdom from his dark magic.": {
(*@ \textcolor{blue}{Plot:Ending} $\triangleright$ @*)          "The protagonist, after facing numerous challenges and                     battles, finally retrieves the powerful artifact.                     With the artifact in hand, they confront the antagonist                     and engage in an epic battle. In the end, the protagonist                     emerges victorious, banishing the sorcerer's dark magic                     and restoring peace to the kingdom of Eldoria.": {
(*@ \textcolor{blue}{Plot:Twist} $\triangleright$ @*)              "Twist: The powerful artifact turns out to have a mind of                       its own and challenges the protagonist, forcing them to                       prove their worth in a final test of courage and                        integrity before it will unleash its true power against                       the antagonist.":
(*@ Identifier $\triangleright$ @*)                 "05e32656-a3b2-47e5-9a81-e6dd312efe33"
          }
        }
      }
    }
  }
}
\end{lstlisting} \\
        \bottomrule
    \end{tabular}
    \caption{Manually pre-defined 14 theme candidates and a premise design within the collected nested dictionary.}
    \label{tab:examples_of_design_of_premise}
\end{table*}

\onecolumn
\begin{small}
\begin{longtable}{>{\raggedright\arraybackslash\tt}p{0.98\textwidth}<{}}
\toprule
\keyword{\textbf{Premise}}: A powerful Roman general, granted immortality by a vengeful deity, must choose between loyalty to the empire and leading a rebellion to bring justice and equality to the corrupt rulers of the Roman Empire. \\
\midrule
\keyword{\textbf{Id}}: fb3af897-3e80-47f1-b632-088640dd40c1
\\
\midrule
\keyword{\textbf{The script generated by Dramatron}}:\\
\\
Title: \textbf{The Immortal Gladiator: A Rebellion for Justice} \\
\\
====\\
\\
The script is based on the storyline:
\textbf{A powerful Roman general, granted immortality by a vengeful deity, must choose between loyalty to the empire and leading a rebellion to bring justice and equality to the corrupt rulers of the Roman Empire.} \\
\\
====\\
\\
\textbf{Maximus Decimus Meridius}: Maximus Decimus Meridius is the powerful Roman general. He is granted immortality by a vengeful deity and must choose between loyalty to the empire and leading a rebellion to bring justice and equality to the corrupt rulers of the Roman Empire. \\
\\
\textbf{Commodus}: Commodus is the corrupt ruler of the Roman Empire. He is the antagonist and a threat to Maximus and his rebellion.\\
\\
\textbf{Livius}: Livius is a loyal friend and ally of Maximus. He supports Maximus in his quest for justice and equality.\\
\\
\textbf{Aurelia}: Aurelia is a strong and independent woman who becomes a key figure in Maximus's rebellion. She is determined to fight for a better future for the people of Rome.\\
\\
\textbf{Lucius}: Lucius is a young and impressionable boy who looks up to Maximus as a father figure. He becomes a symbol of hope for the rebellion.\\
\\
====\\
\\
\textbf{Scene 1} \\
Place: Roman Empire. \\
Plot element: The Ordinary World \\ 
Beat: Maximus is a respected and powerful general in the Roman army, loyal to the empire and its rulers. \\
\\
\textbf{Scene 2} \\
Place: Roman Colosseum. \\
Plot element: Call to Adventure\\
Beat: Maximus is betrayed by Commodus and sentenced to death. He is granted immortality by a vengeful deity and is given a choice to either seek revenge or bring justice and equality to the corrupt rulers of the Roman Empire.\\
\\
\textbf{Scene 3}\\
Place: Roman Senate.\\
Plot element: Refusal of the Call\\
Beat: Maximus initially refuses to lead a rebellion, feeling torn between his loyalty to the empire and his desire for justice.\\
\\
\textbf{Scene 4}\\
Place: Roman Army Camp.\\
Plot element: Crossing the First Threshold\\
Beat: After witnessing the suffering of the people under the rule of Commodus, Maximus decides to lead a rebellion and fight for justice and equality.\\
\\
\textbf{Scene 5}\\
Place: Roman City Streets.\\
Plot element: Tests, Allies, and Enemies\\
Beat: Maximus gathers allies, including his loyal friend Livius, the strong and independent Aurelia, and the young and impressionable Lucius. They face opposition from the corrupt rulers and their loyal followers.\\
\\
\textbf{Scene 6}\\
Place: Roman Palace.\\
Plot element: The Approach to the Inmost Cave\\
Beat: Maximus and his allies plan their rebellion, strategizing how to overthrow the corrupt rulers and bring about change in the Roman Empire.\\
\\
\textbf{Scene 7}\\
Place: Roman Colosseum.\\
Plot element: The Ordeal\\
Beat: The rebellion faces a major setback as Commodus and his forces launch a brutal attack. Maximus and his allies must fight for their lives and the future of the rebellion.\\
\\
\textbf{Scene 8}\\
Place: Roman Senate.\\
Plot element: The Reward\\
Beat: Despite the losses, the rebellion gains momentum and support from the people. They are one step closer to achieving their goal of justice and equality.\\
\\
\textbf{Scene 9}\\
Place: Roman City Streets.\\
Plot element: The Road Back\\
Beat: Maximus and his allies prepare for the final battle against Commodus and the corrupt rulers, knowing that the fate of the Roman Empire rests on their shoulders.\\
\\
\textbf{Scene 10}\\
Place: Roman Colosseum.\\
Plot element: The Resurrection\\
Beat: In a climactic battle, Maximus confronts Commodus and the corrupt rulers, leading the rebellion to victory and bringing about the change they fought for.\\
\\
\textbf{Scene 11}\\
Place: Roman Senate.\\
Plot element: The Return\\
Beat: Maximus and his allies are hailed as heroes, and the Roman Empire begins a new era of justice and equality under their leadership.\\
\\
====\\
\textbf{INT/EXT. Roman Empire. - Scene 1}\\
\\
MAXIMUS\\
I have served the empire faithfully for many years, but I can no longer turn a blind eye to the corruption and injustice that plagues our rulers.\\
\\
SENATOR\\
You speak treason, Maximus. The emperor will not tolerate such talk.\\
\\
MAXIMUS\\
I speak the truth, senator. The people suffer while the rulers grow fat on their wealth and power.\\
\\
SENATOR\\
You would betray the empire and lead a rebellion?\\
\\
MAXIMUS\\
I would lead a rebellion for the people, for justice and equality. I cannot stand by and watch as the empire crumbles under the weight of its own greed.\\
\\
SENATOR\\
You are a fool, Maximus. The emperor will crush any rebellion with the full force of the Roman army.\\
\\
MAXIMUS\\
Perhaps. But I will not stand idly by while the people suffer. I will fight for their freedom, even if it means defying the empire itself.\\
\\
SENATOR\\
You will regret this, Maximus. The emperor will not forgive such betrayal.\\
\\
MAXIMUS\\
I am prepared to face the consequences of my actions. The people deserve a leader who will fight for their rights, and I will be that leader.\\
\\
SENATOR\\
You are a brave man, Maximus. But remember, the consequences of rebellion are severe. Are you willing to risk everything for the sake of the people?\\
\\
MAXIMUS\\
I am. I cannot ignore the suffering of the people any longer. I will do whatever it takes to bring justice and equality to the empire, even if it means facing the wrath of the emperor.\\
\\
SENATOR\\
Very well, Maximus. You have made your choice. But know that the path you have chosen is a dangerous one, and the consequences will be grave.\\
\\
MAXIMUS\\
I am prepared for whatever may come. The people deserve a leader who will fight for their freedom, and I will be that leader, no matter the cost.\\
\\
\textbf{INT/EXT. Roman Colosseum. - Scene 2}\\
\\
Commodus is the corrupt ruler of the Roman Empire. He is the antagonist and a threat to Maximus and his rebellion.\\
\\
COMMODUS\\
Maximus, you have been a loyal and powerful general in my army. But your loyalty has become a threat to my rule. You have been sentenced to death for treason.\\
\\
MAXIMUS\\
I have served the empire faithfully, Commodus. I have fought for justice and honor.\\
\\
COMMODUS\\
Your sense of justice is a threat to my power. But I offer you a chance for redemption. Join me, and together we can rule the empire as father and son.\\
\\
MAXIMUS\\
I will never join you, Commodus. You have betrayed the people and the ideals of Rome. I will not stand by and watch as you continue to oppress and corrupt the empire.\\
\\
COMMODUS\\
Then you leave me no choice. You will face death in the Colosseum, and the people will witness the consequences of defying me.\\
\\
MAXIMUS\\
I will not die, Commodus. I have been granted immortality by a vengeful deity. I will use this gift to bring justice and equality to the corrupt rulers of the Roman Empire.\\
\\
COMMODUS\\
You are a fool, Maximus. Immortality will not save you from my wrath. But if you choose to defy me, then prepare for the consequences. The people will see who truly holds the power in Rome.\\
\\
MAXIMUS\\
I am not afraid of your threats, Commodus. I will lead a rebellion to bring an end to your tyranny and restore honor to Rome. The people will rise up against you, and justice will prevail.\\
\\
COMMODUS\\
We shall see, Maximus. But remember, the consequences of defying me will be severe. You have been warned.\\
\\
MAXIMUS\\
I am prepared to face whatever consequences come my way. I will not rest until the corrupt rulers of the Roman Empire are brought to justice. This is my call to adventure, and I will answer it with courage and determination.\\
\\
\\
\textbf{INT/EXT. Roman Senate. - Scene 3}\\
\\
MAXIMUS\\
I cannot do this. I cannot betray the empire that I have served for so long.\\
\\
SENATOR\\
But Maximus, the empire is corrupt. The rulers are unjust and the people suffer under their tyranny. You have the power to change that.\\
\\
MAXIMUS\\
I understand that, but I cannot simply turn against everything I have fought for. I have sworn my loyalty to Rome, and I cannot break that oath.\\
\\
SENATOR\\
But what about the oath to justice and equality? What about the suffering of the people?\\
\\
MAXIMUS\\
I cannot deny that it weighs heavily on my heart. But I cannot make this decision lightly. I need time to consider my options.\\
\\
SENATOR\\
Time is a luxury we do not have, Maximus. The people need a leader, someone they can rally behind to bring about change.\\
\\
MAXIMUS\\
I know, but I cannot be that leader. Not yet, at least.\\
\\
SENATOR\\
Then when, Maximus? When will you be ready to take a stand for what is right?\\
\\
MAXIMUS\\
I do not know. But I will not be rushed into making a decision that could have dire consequences for the empire.\\
\\
SENATOR\\
Very well, Maximus. But know that the people will not wait forever for a savior. The time will come when they will rise up with or without you.\\
\\
MAXIMUS\\
I understand. And I will not stand in their way. But for now, I must refuse the call to rebellion.\\
\\
SENATOR\\
Very well, Maximus. We will respect your decision for now. But know that the fate of the empire rests in your hands. We can only hope that you will make the right choice when the time comes.\\
\\
\textbf{INT/EXT. Roman Army Camp. - Scene 4}\\
\\
COMMODUS\\
Maximus, you have been a loyal and trusted general in my army. I trust that you will continue to serve the empire with unwavering loyalty.\\
\\
MAXIMUS\\
I cannot, Commodus. I have seen the suffering of the people under your rule. I cannot stand by and watch as injustice and corruption run rampant in the empire.\\
\\
COMMODUS\\
You dare defy me, Maximus? You dare challenge the authority of the emperor?\\
\\
MAXIMUS\\
I do not seek to challenge your authority, Commodus. I seek to bring justice and equality to the people who have suffered under your rule.\\
\\
COMMODUS\\
You are a fool, Maximus. You will regret this decision. The empire will crush any rebellion that dares to rise against it.\\
\\
MAXIMUS\\
I am willing to face the consequences, Commodus. I will not stand idly by while the people suffer. I will lead a rebellion and fight for what is right.\\
\\
COMMODUS\\
You will pay for your betrayal, Maximus. I will not tolerate this defiance.\\
\\
MAXIMUS\\
So be it, Commodus. I am prepared to face whatever comes my way in the pursuit of justice.COMMODUS\\
You may think you are noble, Maximus, but you are nothing but a traitor in the eyes of the empire. You will regret crossing this threshold and defying me.\\
\\
MAXIMUS\\
I have no regrets, Commodus. I will fight for the people and for what is right, no matter the cost.\\
\\
COMMODUS\\
We shall see, Maximus. We shall see.\\
\\
\textbf{INT/EXT. Roman City Streets. - Scene 5}\\
\\
Livius is a loyal friend and ally of Maximus. He supports Maximus in his quest for justice and equality.\\
Aurelia is a strong and independent woman who becomes a key figure in Maximus's rebellion. She is determined to fight for a better future for the people of Rome.\\
Lucius is a young and impressionable boy who looks up to Maximus as a father figure. He becomes a symbol of hope for the rebellion.\\
\\
LIVIUS\\
Maximus, we have gathered our allies and we are ready to fight for justice and equality. But we must be cautious, for our enemies are powerful and ruthless.\\
\\
MAXIMUS\\
I know, my friend. But we cannot stand by and watch the people suffer any longer. We must be strong and united in our cause.\\
\\
AURELIA\\
I am with you, Maximus. I will fight alongside you for a better future for Rome.\\
\\
LUCIUS\\
I want to help too, Maximus. I want to make a difference.\\
\\
MAXIMUS\\
You already have, Lucius. Your presence gives us hope and strength. But we must be prepared for the tests and challenges that lie ahead.\\
\\
LIVIUS\\
We will face many enemies, but we will also find allies who share our vision for a just and equal Rome.\\
\\
AURELIA\\
And we will stand together, united in our cause. We will not be divided or discouraged by those who seek to maintain their power at the expense of the people.\\
\\
MAXIMUS\\
Then let us go forth, my friends. We will face the tests and challenges together, and we will emerge victorious in our quest for justice and equality.\\
\\
LUCIUS\\
I believe in you, Maximus. I know we can make a difference and bring about a better future for Rome.\\
\\
LIVIUS\\
We will stand by your side, Maximus, through thick and thin. Together, we will overcome our enemies and bring about the change that Rome so desperately needs.\\
\\
AURELIA\\
I have faith in our cause and in our unity. We will not falter in the face of adversity, and we will fight for what is right until the very end.\\
\\
MAXIMUS\\
Thank you, my friends. With your unwavering support and determination, I know that we can overcome any obstacle that stands in our way. Let us go forth and make history.\\
\\
\textbf{INT/EXT. Roman Palace. - Scene 6}\\
\\
MAXIMUS\\
We have gathered enough support to make our move. The time has come to overthrow the corrupt rulers and bring justice to the people of Rome.\\
\\
LIVIUS\\
But Maximus, are you sure this is the right path? We could face severe consequences if we fail.\\
\\
MAXIMUS\\
I understand the risks, but I cannot stand by and watch the empire crumble under the weight of corruption and greed. We must take action.\\
\\
AURELIA\\
I stand with you, Maximus. The people deserve better than what they have been given. We will fight for their freedom and equality.\\
\\
LUCIUS\\
I want to help too, Maximus. I want to make a difference.\\
\\
MAXIMUS\\
You have all shown great courage and determination. Together, we will lead the rebellion and bring about change in the Roman Empire.\\
\\
LIVIUS\\
I will follow you to the end, my friend. Let us make our move and bring justice to our people.\\
\\
MAXIMUS\\
It will not be easy, but we are prepared to face whatever challenges come our way. The time has come to approach the inmost cave and confront the corrupt rulers. Are you all ready?\\
\\
AURELIA\\
We are ready, Maximus. Let us bring about the change that Rome so desperately needs.\\
\\
LUCIUS\\
I'm ready, Maximus. I want to make my family proud.\\
\\
MAXIMUS\\
Then let us go forth and make history. The fate of Rome rests in our hands.\\
\\
The group of allies, led by Maximus, prepares to make their move and confront the corrupt rulers of the Roman Empire. They are determined to bring about change and justice for the people of Rome, despite the risks and challenges they may face. With their courage and determination, they are ready to approach the inmost cave and take on the powerful forces that stand in their way. The fate of Rome rests in their hands, and they are prepared to fight for the future of their empire.\\
\\
\textbf{INT/EXT. Roman Colosseum. - Scene 7}\\
\\
COMMODUS\\
You thought you could defy me and get away with it, Maximus? You and your pathetic rebellion are nothing but a nuisance to me.\\
\\
MAXIMUS\\
We will not be silenced, Commodus. We fight for justice and equality, for the people of Rome who suffer under your tyranny.\\
\\
COMMODUS\\
Justice and equality? Ha! You are a fool, Maximus. The people of Rome are nothing but pawns to be used and discarded as I see fit.\\
\\
MAXIMUS\\
You may have the power now, but the people will rise against you. They will not stand for your cruelty and oppression forever.\\
\\
COMMODUS\\
The people will do as I command, or they will suffer the consequences. And as for you, Maximus, you will not live to see the end of this rebellion.\\
\\
MAXIMUS\\
I will not be intimidated by you, Commodus. I will fight for the freedom of Rome until my last breath.\\
\\
COMMODUS\\
Then prepare to meet your end, Maximus. Your rebellion ends here, in the blood-soaked sands of the Colosseum.\\
\\
MAXIMUS\\
So be it, Commodus. We will see who emerges victorious in this battle for the future of Rome.\\
COMMODUS\\
You may have the support of the people, but I have the power of the empire behind me. Your rebellion is nothing but a futile attempt to defy the inevitable.\\
\\
MAXIMUS\\
The people will not be swayed by your empty promises and false power, Commodus. They will see through your lies and stand with us in the fight for freedom.\\
\\
COMMODUS\\
Enough talk, Maximus. It's time to settle this once and for all. Prepare to face the might of the empire and meet your fate in the arena.\\
\\
MAXIMUS\\
I am ready, Commodus. Let the people witness the true strength and resilience of those who fight for justice and equality. The rebellion will not be silenced, and the empire will fall to the will of the people.\\
\\
\textbf{INT/EXT. Roman Senate. - Scene 8}\\
\\
MAXIMUS\\
My fellow senators, the time has come for us to take a stand against the corruption and tyranny that has plagued our empire for too long. We have the support of the people, and together we can bring about a new era of justice and equality.\\
\\
SENATOR 1\\
But Maximus, are you not loyal to the empire? Is this not treason?\\
\\
MAXIMUS\\
I am loyal to the ideals of Rome, to the principles of justice and fairness. It is the corrupt rulers who have betrayed the empire, and it is our duty to restore honor and integrity to our government.\\
\\
SENATOR 2\\
But what of Commodus and his forces? They will not take kindly to our rebellion.\\
\\
MAXIMUS\\
We have faced their brutality before, and we will continue to do so until we have achieved our goal. The people are with us, and together we are strong.\\
\\
SENATOR 3\\
And what of your immortality, Maximus? Will you risk it all for this rebellion?\\
\\
MAXIMUS\\
I have been granted immortality by a vengeful deity, but I will not let that gift dictate my actions. I will fight for the future of Rome, for the future of our people, and I will not rest until justice is served.\\
\\
SENATOR 4\\
Then we stand with you, Maximus. We will fight alongside you and bring about the change that our empire so desperately needs.\\
\\
MAXIMUS\\
Thank you, my fellow senators. Together, we will bring about a new era for Rome, one of justice, equality, and honor.\\
\\
SENATOR 5\\
I have always believed in your leadership, Maximus. I will do everything in my power to support this rebellion and bring about the change we so desperately need.\\
\\
MAXIMUS\\
Thank you, my friend. With the support of the people and the determination of the senate, we will overcome any obstacle in our path. The time for change is now, and together we will achieve our reward - a just and fair Roman Empire.\\
\\
\textbf{INT/EXT. Roman City Streets. - Scene 9}\\
\\
MAXIMUS\\
We have come too far to turn back now. The people are counting on us to bring an end to the tyranny of Commodus and his corrupt rule.\\
\\
ALLIES\\
We stand with you, Maximus. We will fight to the end for justice and equality for all.\\
\\
MAXIMUS\\
The road ahead will be difficult, but we must stay united and focused on our goal. We cannot let fear or doubt cloud our judgment.\\
\\
ALLIES\\
We are ready to follow you into battle, Maximus. Together, we will bring an end to the oppression and restore honor to the Roman Empire.\\
\\
MAXIMUS\\
Prepare yourselves, my friends. The final battle is upon us. We will not rest until we have achieved victory and brought an end to the reign of Commodus.\\
\\
ALLIES\\
For justice and equality! For the people of Rome!\\
\\
MAXIMUS\\
For the future of the Roman Empire! We will not falter, we will not fail. Our cause is just, and our determination is unwavering. Let us march forward and show Commodus and his corrupt allies that the people will not be oppressed any longer. Victory is within our grasp, and we will seize it with all our might!\\
\\
\textbf{INT/EXT. Roman Colosseum. - Scene 10}\\
\\
MAXIMUS\\
Commodus, your reign of tyranny ends here. The people have risen up against you, and justice will be served.\\
\\
COMMODUS\\
You dare challenge me, Maximus? I am the ruler of the Roman Empire, and I will not be overthrown by a mere mortal like you.\\
\\
MAXIMUS\\
You may have the title of emperor, but you have lost the respect and loyalty of the people. Your corrupt ways have brought suffering and injustice to the empire, and it is time for a new era of equality and justice.\\
\\
COMMODUS\\
You speak of justice, yet you lead a rebellion against the rightful ruler of Rome. You are a traitor and a threat to the stability of the empire.\\
\\
MAXIMUS\\
I am no traitor. I fight for the people, for their freedom and their rights. You have abused your power and brought ruin to the empire. It is time for a new beginning, a resurrection of the values that Rome was built upon.\\
\\
COMMODUS\\
You will not succeed, Maximus. I will crush your rebellion and maintain my grip on power. The people will learn to fear and obey me once more.\\
\\
MAXIMUS\\
Your reign of terror ends now, Commodus. The people have chosen to stand with me, and together we will bring about the resurrection of justice and equality in the Roman Empire.\\
\\
(COMMODUS and MAXIMUS engage in a fierce battle, with the fate of the empire hanging in the balance. In the end, MAXIMUS emerges victorious, and the people celebrate the resurrection of a new era in Rome.)\\
\\
\textbf{INT/EXT. Roman Senate. - Scene 11}\\
\\
MAXIMUS\\
My fellow senators, I stand before you today not as a conqueror, but as a servant of the people. The time has come for us to rebuild our empire, not on the backs of slaves and corruption, but on the principles of justice and equality.\\
\\
SENATOR 1\\
But Maximus, you were once a loyal general of the empire. How can we trust that you will not lead us astray?\\
\\
MAXIMUS\\
I understand your concerns, but I have been granted immortality by the gods themselves. I have seen the rise and fall of empires, and I have learned from the mistakes of the past. I will not let power corrupt me, and I will always put the needs of the people first.\\
\\
SENATOR 2\\
But what of the emperor? Will you not be seen as a traitor to the throne?\\
\\
MAXIMUS\\
The emperor has lost his way, and it is time for a new leader to rise. I do not seek power for myself, but for the betterment of our empire. I will lead with honor and integrity, and I will ensure that the voices of the people are heard.\\
\\
SENATOR 3\\
It is a bold vision, Maximus. But can we truly trust that you will bring about the change you promise?\\
\\
MAXIMUS\\
I cannot force you to trust me, but I ask that you look at my actions and the actions of my allies. We have fought for justice and equality, and we will continue to do so. I will not rest until the Roman Empire is a place where every citizen can thrive and live in peace.\\
\\
SENATOR 4\\
We will support you, Maximus. We have seen the change you have brought about, and we believe in your vision for the future of our empire.\\
\\
MAXIMUS\\
Thank you, my friends. Together, we will build a new Rome, a Rome that stands as a beacon of hope and justice for all.\\
\\
SENATOR 5\\
I pledge my loyalty to you, Maximus. Let us work together to create a better future for our people.\\
\\
MAXIMUS\\
Thank you, my fellow senators. With your support, we will bring about the change that our empire so desperately needs. Together, we will build a Rome that we can all be proud of.\\
\\
====\\
\midrule 
\keyword{\textbf{The novel generated by RecurrentGPT}}:\\
\\
Name:\textbf{ Immortal Legion}\\
\\
Outline:\\
\\
- \textbf{Introduce the protagonist, Marcus Aurelius, a revered Roman general known for his unwavering loyalty to the empire.}\\
\\
- \textbf{Describe the pivotal battle where Marcus is granted immortality by a vengeful deity, cursing him to live forever until he chooses to defy the empire.}\\
\\
-\textbf{ Set the stage for the internal conflict Marcus faces as he grapples with his newfound immortality and the corrupt nature of the Roman rulers.}\\
\\
The sun blazed fiercely over the battlefield, casting a harsh glare upon the clashing armies. Marcus Aurelius, a formidable Roman general, stood at the forefront of his legion, his eyes ablaze with determination. The air was thick with the scent of blood and sweat, and the deafening clash of swords echoed through the valley. Marcus had always been a loyal servant of the empire, leading his men with unwavering devotion. But as the battle raged on, a sense of foreboding crept over him, as if the fates themselves were preparing to intervene.\\
\\
In the midst of the chaos, a sudden bolt of lightning split the sky, striking the ground mere feet from where Marcus stood. As the blinding light faded, a figure emerged from the smoke, emanating an otherworldly aura. It was a vengeful deity, angered by the empire's tyranny and corruption. In a voice that resonated with power, the deity cursed Marcus with immortality, binding him to an existence that would endure until he chose to defy the empire and seek justice for its oppressed citizens. As the deity vanished, leaving Marcus bewildered and burdened with a weighty choice, the battle around him reached its brutal climax.\\
\\
In the aftermath of the battle, Marcus found himself grappling with the implications of his newfound immortality. He had always been a loyal servant of Rome, but now, the deity's curse had planted a seed of doubt within him. The empire's corruption and oppression weighed heavily on his conscience, and he knew that he could no longer turn a blind eye to the suffering of its people. As he gazed upon the bloodstained battlefield, a steely resolve took hold in his heart. Marcus Aurelius, the immortal general, would rise to challenge the very empire he had once served, in a quest to bring justice and equality to the citizens who had long suffered under its rule.\\
\\
With a heavy heart and a resolute mind, Marcus Aurelius embarked on his journey to challenge the empire. He sought out like-minded individuals who shared his vision of justice and equality, carefully weaving a network of allies while concealing the truth of his immortality. The task of uniting disparate factions proved to be a daunting challenge, as internal conflicts threatened to tear his rebellion apart. As he navigated the treacherous landscape of political intrigue and deception, Marcus grappled with the weight of his newfound purpose, wrestling with conflicting emotions that threatened to consume him. The internal and external conflicts he encountered tested the limits of his resolve and shaped the path of his rebellion, setting the stage for a battle that would determine the fate of the empire.\\
\\
As Marcus and Livia continued to forge their rebellion, they encountered a group of skilled and diverse individuals who shared their passion for justice. Among them was Lucius, a former scholar turned warrior, who brought a wealth of knowledge and insight to their cause. His strategic mind and ability to navigate the intricate web of political alliances proved invaluable. Alongside Lucius was Ariadne, a fierce and charismatic leader who commanded the respect and loyalty of her followers. Her unwavering determination and eloquence inspired hope and unity among the rebellion's ranks. As the group expanded, they faced increasing resistance from the empire, leading to daring escapades and narrow escapes. Despite the challenges, their bond strengthened, fueled by a shared vision of a liberated and just society. Together, they strategized and planned, each member bringing their unique skills and experiences to the table, shaping the rebellion into a formidable force that could no longer be ignored by the empire.\\
\\
The rebellion faced increasing resistance from the empire, forcing Marcus, Livia, Lucius, and Ariadne to navigate treacherous situations. As they sought to expand their network of supporters, they encountered a mysterious figure who presented them with a tantalizing opportunity to shift the balance of power in their favor. This influential individual offered a chance to gain crucial information about the empire's vulnerabilities, potentially tipping the scales in their favor. However, the conditions of this alliance came with a perilous price, one that could compromise the essence of their rebellion and sow seeds of discord within their ranks. The offer presented a moral dilemma, testing the loyalty and convictions of each member and setting the stage for internal conflict and external consequences. As they grappled with this decision, Marcus and his allies found themselves embroiled in a web of intrigue and danger, where trust and betrayal intertwined, threatening to unravel all they had worked for.\\
\\
The high-ranking official's words hung in the air, each syllable laden with the weight of their potential consequences. Marcus and his allies listened intently, their minds racing with the implications of the information being offered. The official's promises of crucial intelligence were like a siren's call, beckoning them to grasp at the chance to weaken the empire's grip on power. Yet, as they weighed the risks and rewards, a sense of unease settled over the group, each member grappling with their own fears and doubts. The tension in the chamber was palpable, the air thick with the conflicting desires for liberation and the fear of betrayal. Despite the allure of the official's offer, a shadow of doubt lingered in Marcus's mind, a nagging suspicion that this tempting opportunity was too good to be true. As the meeting drew to a close, the group left the opulent chamber, their thoughts consumed by the precarious path that lay ahead. They knew that their next steps would shape the course of their rebellion, and the consequences of their choices would echo through history, for better or for worse.\\
\\
As the weight of doubt and uncertainty continued to gnaw at Marcus, he sought counsel from an enigmatic figure with ties to the empire, hoping to unravel the true intentions behind the official's offer. The clandestine meetings, shrouded in secrecy and danger, raised suspicions and drew the attention of those who would see the rebellion crushed. Marcus found himself navigating a treacherous web of deceit and betrayal, his every move scrutinized by shadowy figures within the empire's ranks. Despite the looming threat of discovery, Marcus was determined to unearth the truth and safeguard the rebellion's future, even if it meant delving into the heart of the empire's dark underbelly.\\
\\
The clandestine meetings with Xue continued to unravel layers of intrigue and danger, drawing Marcus deeper into a web of deception and shifting loyalties. As Marcus probed for clues and sought to untangle the truth behind the official's offer, he found himself entangled in a delicate game of cat and mouse with the empire's clandestine operatives. His pursuit of the truth led to unexpected alliances and dangerous revelations, plunging the rebellion into an even more precarious position. The enigmatic figure, Xue, held the key to unlocking the empire's dark underbelly, but his allegiances remained veiled in mystery, raising questions of trust and betrayal. With powerful adversaries within the empire taking notice of Marcus's actions, the conflict escalated, setting the stage for a high-stakes confrontation that would test the rebellion's resilience and the strength of their convictions.\\
\\
The plan for a daring operation to confront the empire's most feared enforcer began to take shape as Marcus and his allies gathered in the dimly lit chamber of their secret meeting place. Tension crackled in the air as they pored over maps and whispered urgently, plotting their risky maneuver to gather vital intelligence. The weight of uncertainty hung heavy in the room, but determination burned in Marcus's eyes as he outlined the details of their audacious plan. Each member of the rebellion knew the perilous stakes and the potential consequences of their actions, yet they stood resolute in their resolve to challenge the empire's iron grip. As the hours slipped away, the final preparations fell into place, and a sense of grim anticipation settled over the group, mingled with the hope of a decisive victory that could shift the tides of their rebellion.\\
\\
The enigmatic figure, Xue, had led Marcus deeper into the treacherous underworld of the city, where the shadows whispered of secrets and danger. Each clandestine encounter with Xue unraveled new layers of complexity, revealing a world where loyalties were as transient as the shifting sands of the desert. Despite the looming threat of discovery, Marcus was determined to unearth the truth and safeguard the rebellion's future, even if it meant delving into the heart of the empire's dark underbelly. As the weight of uncertainty continued to gnaw at him, Marcus found himself entangled in a delicate game of cat and mouse with the empire's clandestine operatives. The stakes soared to dizzying heights as he delved into the heart of the empire's dark underbelly, where danger lurked in every shadow and betrayal loomed like a specter. Yet, with each perilous step, Marcus unearthed fragments of truth that painted a chilling portrait of the empire's inner workings, a revelation that would reshape the rebellion's path and plunge them into an even more precarious position.\\
\\
As the urgency of the rebellion's cause drives Marcus forward, he embarks on a perilous mission to infiltrate a heavily guarded facility rumored to hold vital intelligence crucial to the rebellion's success. The labyrinthine corridors of the facility present formidable obstacles, and the ever-present threat of discovery looms over him, heightening the stakes to unprecedented levels. With the weight of responsibility pressing down on him, Marcus navigates the treacherous passages, his every move calculated and deliberate to ensure the safety of his allies and the future of the rebellion. The shadows of the facility whisper of secrets and danger, and Marcus finds himself entangled in a high-stakes game of deception and survival. The pulse of the city beats with an undercurrent of tension and unrest, fueling Marcus's determination to unearth the truth, no matter the cost. As he inches closer to the heart of the facility, the web of deceit and betrayal tightens around him, and the looming confrontation with the empire's most feared enforcer threatens to plunge him into a perilous game of brinksmanship and deception.\\
\\
As Marcus cautiously navigates the labyrinthine corridors of the heavily guarded facility, he stumbles upon a clandestine chamber hidden deep within the heart of the structure. To his astonishment, the chamber is filled with encrypted documents that unravel a web of deceit and betrayal, shedding light on the empire's most closely guarded secrets. As he sifts through the cryptic messages, a chilling realization dawns upon him – the empire's most trusted allies may harbor treacherous intentions, and the rebellion's very existence hangs in a precarious balance. With each revelation, Marcus finds himself plunged into a harrowing game of survival, where every decision could mean the difference between victory and defeat. The weight of responsibility presses down on him, and the shadows of the facility seem to whisper of imminent danger, heightening the stakes to unprecedented levels. As he grapples with the shocking truths laid bare before him, Marcus is forced to confront the harsh reality that the path to victory may be paved with sacrifice and betrayal.\\
\\
\bottomrule
\caption{A example of script and novel generated from a MoPS premise by Dramatron and RecurrentGPT.}
\label{tab:script_and_novel}
\end{longtable}
\end{small}

\end{document}